\documentclass{article}

\usepackage[final]{corl_2018} 

\usepackage{microtype}
\usepackage{graphicx}
\usepackage{booktabs} 
\usepackage{hyperref}
\usepackage{url}            
\usepackage{amsfonts}       
\usepackage{nicefrac}       
\usepackage{microtype}      
\usepackage[utf8]{inputenc}
\usepackage{amsmath}
\makeatletter
\g@addto@macro\normalsize{%
  \setlength\abovedisplayskip{4pt}
  \setlength\belowdisplayskip{4pt}
  \setlength\abovedisplayshortskip{4pt}
  \setlength\belowdisplayshortskip{4pt}
}
\usepackage{bbm}  
\usepackage{graphicx}
\usepackage{wrapfig}
\usepackage{float}
\usepackage[font={small}]{caption}
\usepackage{mathtools}
\usepackage{color}
\usepackage{verbatim}
\usepackage{tikz}
\usepackage{enumitem}  
\usepackage{cancel}
\usepackage{times} 
\usepackage{amssymb,longtable,calc}
\usepackage{mathptmx}
\usepackage{xspace}
\usepackage{latexsym}
\usepackage{multirow}
\usepackage{color,comment,rotating,url,xspace} 
\usepackage{subcaption}
\usepackage{tabularx}
\usepackage[ruled,vlined]{algorithm2e}  
\newcolumntype{Y}{>{\centering\arraybackslash}X}

\makeatletter
\let\@minipagerestore=\raggedright
\makeatother
\newlength{\mylen}
\setbox1=\hbox{$\bullet$}\setbox2=\hbox{\tiny$\bullet$}
\setlist[itemize]{itemsep=0mm, topsep=2pt}

\DeclareMathOperator*{\argmax}{arg\,max}
\DeclareMathOperator*{\diag}{diag}

\def\tabrowsep{\noalign{\vskip 2pt}}

\title{Global Search with Bernoulli Alternation Kernel for Task-oriented Grasping Informed by Simulation}

\author{
  Rika Antonova\thanks{Both of these authors contributed equally.\vspace{-5px}}, $\ \ $ Mia Kokic\footnotemark[1], $\ \ $ Johannes A. Stork, $\ \ $ Danica Kragic  \\
  Robotics, Perception and Learning, CSC \\
  KTH Royal Institute of Technology
  Sweden \\
  \texttt{\{antonova, mkokic, jastork, dani\}@kth.se} \\
}

\begin{document}
\maketitle

\vspace{-15px}
\begin{abstract}
\small{We develop an approach that benefits from large simulated datasets and takes full advantage of the limited online data that is most relevant. We propose a variant of Bayesian optimization that alternates between using informed and uninformed kernels. With this Bernoulli Alternation Kernel we ensure that discrepancies between simulation and reality do not hinder adapting robot control policies online.
The proposed approach is applied to a challenging real-world problem of task-oriented grasping with novel objects. Our further contribution is a neural network architecture and training pipeline that use experience from grasping objects in simulation to learn grasp stability scores. We learn task scores from a labeled dataset with a convolutional network, which is used to construct an informed kernel for our variant of Bayesian optimization.
Experiments on an ABB Yumi robot with real sensor data demonstrate success of our approach, despite the challenge of fulfilling task requirements and high uncertainty over physical properties of objects.}
\end{abstract}
\vspace{-5px}

\vspace{-10px}
\section{Introduction}
\label{sec:intro}
\vspace{-5px}
Recent advances in deep learning motivated using data-driven methods in robotics.
However, collecting large amounts of training data is challenging, since it requires actual execution on a real system. One way to trim hours of execution time is to use simulation. Simulators make simplifying assumptions, so often there is mismatch between simulation and real world. 
We propose to bridge this gap by using a variant of Bayesian optimization (BO) that incorporates simulation-based information in a way that is robust to mismatch between simulation and reality. 
Furthermore, we apply this to the problem of task-oriented grasping.  
The challenge is that successful grasps are limited to specific object parts that afford the execution of a task. This presents a highly constrained problem.

\begin{figure}[t]
\centering
\includegraphics[width=1.0\textwidth]{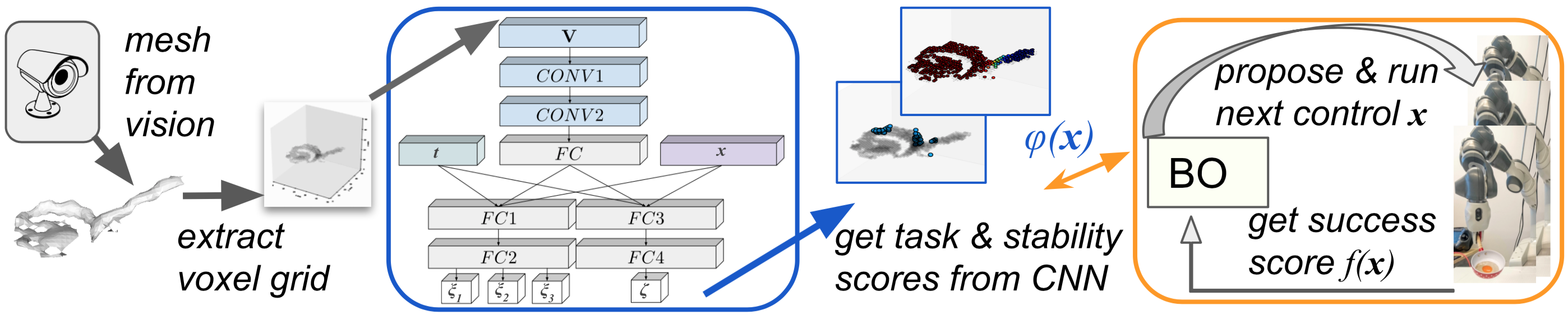}
\caption{\small{Overview of the overall approach. Blue highlights components trained offline; orange highlights online learning components that guide real-time decisions made by the robot. A vision system is used to first construct a mesh from raw point cloud. Then, a set of grasping points, voxelized mesh and task identity are passed through the network to obtain task suitability and grasp stability scores. These are used to construct informed kernel for BO. Then, the robot executes online search with BO to find the best task-appropriate grasp.}}
\label{fig:approach}
\vspace{-10px}
\end{figure}

Our first contribution is a variant of BO that alternates between using informed and uninformed kernels. This allows us to explore grasp adjustments online, while exploiting simulation-based information and feedback from previous grasp attempts. To inform the kernel for our BO algorithm, we propose a task-oriented grasp model. This further contribution is a deep neural network architecture that maps raw visual observation of an object to task-oriented grasp scores. The network is trained on large amounts of synthetic data. This allows us to compute task-oriented grasp candidates for previously unseen objects from online perceptual information. Simulation-based knowledge provides additional guidance for online search, making it more sample-efficient.
Overall, our approach aims to emulate the human-like strategy of attempting several adjustments until a successful grasp is accomplished; Figure~\ref{fig:approach} gives a brief visual overview.
We conduct experiments on an ABB Yumi robot with real sensor data, and demonstrate success of our approach to fulfill task requirements when grasping novel objects, despite high uncertainty over their physical properties.

\vspace{-2px}
\section{Background and Related Work}
\label{sec:background}
\vspace{-5px}

\vspace{-2px}
\subsection{Bayesian Optimization Informed By Simulation}
\label{subsec:background_bo}
\vspace{-5px}

Bayesian optimization (BO) is a data-efficient global search method (see \cite{BOtutorial2016} for an overview). 
Let $\pmb{x}$ be a set of control parameters (e.g. for grasping: the approach direction and orientation of the end-effector). The problem is to find $\pmb{x}$ that optimizes a given objective/cost function $f(\pmb{x})$. The objective function encodes characteristics of the desired outcome: e.g. low score for failing to grasp an object, high score for grasping and completing the desired task.
BO starts with a prior expressing uncertainty over $f(\pmb{x})$. After each real-world trial, BO constructs a posterior based on data obtained so far. It then uses an auxiliary function (\textit{acquisition function}) to choose the next $\pmb{x}$ to evaluate. The acquisition function selects points for which the posterior estimate of the objective $f$ is promising. It takes into account both the posterior mean and covariance. Gaussian Process (GP) is commonly used to model the cost function:
$f(\pmb{x}) \sim GP(\mu(\pmb{x}), k(\pmb{x}, \pmb{x}'))$.
Its kernel function captures similarity between inputs: if $k(\pmb{x}, \pmb{x}')$ is large for $\pmb{x}, \pmb{x}'$, then $f(\pmb{x}$) has high influence on $f(\pmb{x}')$. Squared Exponential (SE) kernel is frequently the default choice (its hyperparameters $\sigma_k^2, \ \pmb{\ell}$ can be optimized automatically):
\begin{equation}
\label{eq:k_SE}
k_{SE}(\pmb{x}, \pmb{x}') = \sigma_k^2 \exp\big(- \tfrac{1}{2} (\pmb{x} - \pmb{x}')^T \diag(\pmb{\ell})^{\!-\!2} (\pmb{x} - \pmb{x}') \big),
\end{equation}

BO with standard kernels (SE or, more generally, Mat\'ern kernels) is effective in cases with low noise, smooth objective/cost functions or limited search space. In contrast, robotics problems exhibit high noise, sharp transitions between low- and high-cost regions, inability to limit the search space a priori without excluding optimal regions. Previous work explored using kernels informed by simulation~\cite{cully2015robots,rai2018bayesian} and by the domain dynamics~\cite{marco2017lqr}, in some cases allowing to learn kernel transforms automatically~\cite{antonova2017deep}. However, these assumed access to high-fidelity simulators/models. For many areas of robotics such simulators are not available. 
High anticipated simulation-reality mismatch prevents us from using simulation-based information in standard ways, like constructing the prior of the mean function for the Gaussian Process in BO. As noted in~\cite{GPsMLBook}, in practice it can be challenging to specify prior mean effectively. Recent experiments in robotics show that performance of `prior-based' BO can degrade in cases with significant simulation-reality mismatch~\cite{rai2018using}. Hence, embedding simulation-based information into the kernel could be a more robust alternative. 

\vspace{-2px}
\subsection{Task-oriented Grasping}
\vspace{-5px}
Grasping simulators~\cite{diankov2008openrave} can help generating a set of grasps labeled with quality measures, which can be used to select the best grasp for execution on a real robot. However, these quality measures often rely on access to full 3D geometry of an object, which is often not obtainable in real-world scenarios. To overcome this, recent work investigated learning to predict grasp success from partial visual inputs \cite{lenz2015deep}, \cite{mahler2017dex}, \cite{schmidt2018grasping}. 
Vision alone often does not provide enough information about important object properties: it is difficult to infer inertial and, even more so, friction properties. Hence, several works proposed learning adaptively with feedback from real robot trials \cite{kroemer2010combining, montesano2012active, oberlin2018autonomously}. Some proposed constructing priors from simulation, but as noted in the BO background subsection, directly adding prior points or using a fixed prior mean is problematic with high simulation-reality mismatch. In the case of grasping: only medium- to low-fidelity simulators are generally available; their results only align with reality if crucial parameters, like friction, are measured and modelled meticulously. We do not assume access to estimates of friction or inertial properties of objects, and do not require modeling these in simulation either. 
Our main goal is to develop a method whose performance does not degrade due to incorporating data from highly inaccurate simulations. Hence, we draw inspiration from approaches that embed simulation-based information into the kernel (e.g.~\cite{mahler2016dex}, though this prior work investigated convergence of MABs with a large number of pick-and-place simulated grasps across objects, while we emphasize finding successful task-oriented grasps with only a few BO trials on the real robot and adapt to each object separately).

Approaches from prior work described above can be successful for grasping an object in a stable manner.
However, in most scenarios the ultimate goal is to allow an execution of a manipulation task, for example cutting or pouring. This problem is known as Task-oriented Grasping (TOG) \cite{song2015task}, \cite{antanas2014high} and is vastly more challenging than just stable grasping. The first challenge is to find grasps that are stable and also task compliant. For example, when grasping a mug most of the stable grasps for smaller parallel grippers are on the rim. However, if we want to use the mug for pouring, fingers should avoid the opening area and instead aim for the handle. This is challenging when vision is imperfect, since small task-relevant areas of the object could be obscured or distorted. The second challenge is that a robot must be able to reason about geometric properties of an object and how they relate to tasks (e.g. openings \& pouring, blades \& cutting). This is know as \emph{affordance} learning and many authors have addressed this problem in the past. While some works focus solely on reasoning about object affordances, others utilize them for purpose of TOG. In \cite{kokic2017affordance} authors trained a CNN to predict part affordances, which are used to formulate task constraints, namely the ones on the location and orientation of the gripper. These constraints are given to optimization-based grasp planner, which executes task-oriented grasps. Similarly, \cite{detry2017taskoriented} trained task-oriented CNNs to identify the regions a robot is allowed to contact to fulfill a task; they use a part-based approach that finds object parts with shape that is compatible with the gripper.
In \cite{fang2018learning} authors proposed training TOG Network to optimize for task-oriented grasp and manipulation policy. They decomposed the problem into 1) finding task-agnostic grasps for which they use Dex-Net 2.0~\cite{mahler2017dex} and 2) finding task-oriented grasps for which they train a CNN. Our work differs from this in several ways. We train a network that jointly predicts grasp stability and task suitability. Furthermore, although we could use a single score (like Dex-Net 2.0) for predicting stability, we chose to incorporate three different metrics. This provides a richer signal for BO: different metrics could be of varying importance for different tasks (e.g. one metric could score grasps based on the distance from the center of mass of the object, while another might be most sensitive to positions of the finger joints). Moreover, we train our network on a large data set of realistic 3D objects and six different tasks, while \cite{fang2018learning} use procedurally generated objects based on shape primitives and consider only two tasks (pounding and sweeping). Hence, in our case the complexity of learning object-task-grasp relationships is significantly higher.

\vspace{-8px}
\section{Proposed Method}
\label{sec:methodology}
\vspace{-4px}
\vspace{-2px}
\subsection{Bayesian Optimization with Bernoulli Alternation Kernel}
\label{subsec:approach_bo}
\vspace{-5px}
Our aim is to develop an approach robust to high simulation-reality mismatch. This motivates us to look beyond solutions that put simulation-based information into GP prior or rely on simulation-based kernels alone. We first note that using a sum of kernels could be beneficial. Let $k_{sum}(\pmb{x},\pmb{x}') = k_{SE}(\pmb{x}, \pmb{x}') + k_{\phi}(\pmb{x}, \pmb{x}')$ be a kernel comprised of $k_{SE}$ (Equation~\ref{eq:k_SE}) and $k_{\phi}=k_{SE}(\phi(\pmb{x}), \phi(\pmb{x}'))$, where $\phi(\cdot)$ is akin to a warping function.
Recall that in our case $\pmb{x}$ is a vector of control parameters. 
To obtain $\phi(\pmb{x})$ we could execute controls $\pmb{x}$ in simulation and output relevant characteristics of the result (e.g. stability metrics for a grasp). It is useful to embed $\phi$ into the kernel, because we can collapse the space of unsuccessful controls. For example, $\phi$ could give near zero stability scores to grasps that miss or barely touch objects. This would indicate that all such failed points/controls are similar to each other, but dissimilar from successful regions of control parameters. BO can then quickly learn to neglect the non-promising regions, even though they could be far away from each other in the original space of control parameters. 
However, when $\phi$ fails to provide high-quality information, $k_{sum}$ could be adversely impacted by the $k_{\phi}$ component.

To offer a more robust alternative, we propose an approach that takes contributions from both $k_{SE}$ and $k_{\phi}$, but ensures BO can not be mislead by a poor choice of $\phi$. At each BO iteration/trial, we randomize the choice of whether $k_{SE}$ or $k_{\phi}$ is used. For this, we define a probability distribution over kernels and draw a kernel function to be used at each BO trial independently. This defines $k_{bak}$ as:
\begin{equation}
\label{eq:bak_kernel}
k_{bak}(\pmb{x}, \pmb{x}') \sim \mathbbm{1}_{\{\theta \leq 0.5\}} k_{SE}(\pmb{x}, \pmb{x}') + \mathbbm{1}_{\{\theta > 0.5\}} k_{\phi}(\pmb{x}, \pmb{x}'); \ \  \theta \sim Uniform(0,1)
\end{equation}
Note that after each iteration/trial $n$, the data for computing the GP posterior consists of all the points sampled so far: $\pmb{D}_n = \{ (\pmb{x}_i, f(\pmb{x}_i)) | i=1,...,n \}$. In BO, posterior is usually re-computed after each new sample. 
This aspect of BO allows us to make a choice of the kernel function for each iteration separately. So, after $n$ iterations/trials, we first pick a kernel function $k_n$ using Equation~\ref{eq:bak_kernel}. We then compute GP posterior mean and covariance (notation from \cite{GPsMLBook}):
\begin{align}
mean(f_*) = \pmb{k}_{*_n}^T (K_n+\sigma^2_{noise} I)^{-1} \pmb{y}_n \quad \quad
cov(f_*) = k_n(\pmb{x}_*,\pmb{x}_*) - \pmb{k}_{*_n}^T (K_n+\sigma_{noise}^2 I)^{-1} \pmb{k}_{*_n}
\label{eq:posterior}
\end{align}
where $\pmb{x}_*$ is a new point whose cost we want to predict; $f_* \!:=\! f(\pmb{x}_*)$; $K_n$ is $n \!\times\! n$ matrix with $K_{ij}\!=\!k_n(\pmb{x}_i,\pmb{x}_j)$; $\pmb{k}_{*_n}\!\!\in\! \mathbb{R}^n$ is a vector of covariances between $\pmb{x}_*$ and each $\pmb{x}_{i}, \ i\text{=}1,...,n$; $\pmb{y}_n$ is a vector of evaluations for the sampled points: $[\pmb{y}_n]_i = f(\pmb{x}_i)$. Algorithm~\ref{alg:bo_bak} gives a concise summary. 

\begin{wrapfigure}{r}{0.54\textwidth}
\vspace{-2px}
\begin{algorithm}[H]
  \LinesNumberedHidden
  \DontPrintSemicolon
  sample $\pmb{x}_1$ randomly, get $y_1 = f(\pmb{x}_1)$ from real world\;
  initialize: $\pmb{D}_1 = \{ (\pmb{x}_1, y_1) \}$\;
  \For{$n = 1,2, ... $}{
     sample kernel function $k_n$ using Equation~\ref{eq:bak_kernel}\;
     get posterior GP mean and cov using $\pmb{D}_n$ \& Eq.~\ref{eq:posterior}\;
     select $\pmb{x}_{n+1}$ by optimizing acquisition function $\alpha$:\;
     $\quad \quad \quad \pmb{x}_{n+1} = \argmax_{\pmb{x}} \alpha(\pmb{x}; \pmb{D}_n)$\;
     get $y_{n+1} = f(\pmb{x}_{n+1})$ from real world\;
     augment data $D_{n+1} = \pmb{D}_n \cup \{(\pmb{x}_{n+1}, y_{n+1})\}$
  } 
  \caption{BO-BAK}
  \label{alg:bo_bak}
\end{algorithm}
\vspace{-10px}
\end{wrapfigure}

For intuitive insight, note that points sampled for trials when $k_{\phi}$ is used could provide fast guidance towards useful parts of the search space. These points are included in the data for subsequent posterior computations, enabling even trials in which $k_{SE}$ is used to propose better next choices.
This is important if probability of discovering a promising region is small, which is frequent in robotics. If $k_{\phi}$ is not useful, or even misleading, choices made on the trials that use $k_{SE}$ are not impacted by poor $\phi$, since with our approach BO's acquisition function makes its decisions using only one of the kernels at a time.

In Section~\ref{sec:bo_analytic_tests} we evaluate on analytic functions that exhibit challenges similar to our target domain. We confirm that informed kernels can provide significant improvement over conventional BO. $k_{\phi}$ and $k_{sum}$ appear somewhat sensitive to the quality of the warping function $\phi$, while $k_{bak}$ appears to be more robust empirically. As a preliminary theoretical comment, we observe that $k_{bak}$ retains optimality guarantees of the conventional BO. Intuitively, this is because in expectation we perform N/2 trials using only SE to choose the next point. This is no worse than using conventional BO in half of the trials, and we anticipate that satisfactory consistency and regret bounds could be derived. We leave this theoretical analysis to future work. In this work, we investigate whether $k_{bak}$ is beneficial in complex real-world robotics scenarios. These pose a further challenge, since often the assumptions made to obtain consistency and regret bounds for conventional BO do not hold in practice.
\vspace{-2px}
\subsection{Task-oriented Grasping, Network Architecture and Training}
\label{sub:TOG}
\vspace{-5px}
The objective of our TOG CNN is to learn to output several grasp stability scores $\xi_i$ and a task suitability score $\zeta$, given as input a task $t$ and a grasp $\pmb{x}$ on an object. 
A grasp is parameterized by $\pmb{x}\!:=\!(\pmb{p},\vec{n}, \psi, d)\!\in\!\mathbb{R}^8$, where $\pmb{p}\!\in\! \mathbb{R}^3$ is a point on the surface of the object and $\vec{n}\!\in\!\mathbb{R}^3$ is its corresponding unit normal, $\psi$ is a gripper roll and $d$ is an offset from the surface of the object. A task $t$ is defined as a single manipulation action that starts with a grasp.
To fully exploit the power of 3D representations we use volumetric architecture in convolutional layers. For this, we scale an object mesh to fit inside $50 \times 50 \times 50$ binary voxel grid (for learning, we also scale the grasping points). The actual input to the network is then: a voxel grid of an object ($\pmb{V}$), a grasp ($\pmb{x}$), and a task ($t$, encoded as $1 \times 6$ one-hot vector). For details of network architecture see Figure~\ref{fig:approach}.
To handle noisy data that can be encountered in the real-world, the network is trained with dropout (ratio of $0.5$) in the first layer. Furthermore, for training we randomly rotate objects to account for possible orientations for tabletop grasping. The training was performed using Tensorflow on a Titan~X GPU. 

A single CNN is trained for all object categories and tasks (no need to classify the object explicitly). The kernel for BO is constructed ``on the fly'' with a forward pass on the trained CNN. To grasp a new object the steps are: 1) vision processes  the object (point cloud $\rightarrow$ mesh $\rightarrow$ voxel grid); 2) voxel grid \& task id are fed as input to CNN; 3) we get as output predicted stability \& task scores for any sets of grasp parameters, which enables computing kernel distances quickly during BO. Once the CNN is trained, we have quick access to a kernel for any object type and task that are included in CNN training. BO is run for each object instance ``from scratch'', but with informed kernel incorporated into the search. In collaborative industrial settings (where the same objects could be used for many days) our method could identify optimal task-specific grasping parameters and these can be utilized for a given tool without re-optimizing. If it is unlikely that the robot will deal with the same tool/object again, BO would run each time for each new object instance. This relieves us from making restrictive assumptions about object properties that can't be inferred from vision only.

\vspace{-4px}
\subsection{Data Generation}
\vspace{-2px}
\label{sub:dgs}

To obtain learning targets for our network, we simulate grasping various objects in OpenRave~\cite{diankov2008openrave} with a parallel gripper. We provide the simulator with a ``free-floating'' 3D model of the gripper and objects. The objects dataset consists of $605$ mesh models from ShapeNET~\cite{chang2015shapenet} and ModelNet40~\cite{wu20153d}, containing objects from $13$ different categories. For training we align each mesh with a reference frame that coincides with object's principal axes and meshes are scaled to real-world size based on object's category.

\begin{wrapfigure}{r}{0.58\textwidth}
\vspace{-8px}
{
\centering
\footnotesize
\small\addtolength{\tabcolsep}{-5pt}
\begin{tabular}{c c c}
 \\ \toprule
 Task & Object Class & Grasp Requirements
 \\ \midrule
  handover & all objects & leave handle(s) clear \\ 
  screw & screwdrivers & avoid the shaft  \\ 
  cut & knife and scissors & avoid blade(s) \\ 
  pour & bottle, can, mug, wine glass &  avoid the opening area  \\
  support & pan, spatula, spoon, fork & avoid the supporting area \\
  pound & hammer & avoid a hammer head \\ 
  \bottomrule
\end{tabular}
}
\vspace{-3px}
\caption{\small{Objects, tasks and grasps requirements in our dataset.}}
\label{fig:tasks}
\vspace{-10px}
\end{wrapfigure}

We label the objects with target task scores by assigning positive/negative labels to parts of the object suitable/unsuitable for the task. Figure~\ref{fig:tasks} shows the tasks we consider and relations to applicable objects.
We execute $4500$ grasps on each object with grasp parameters as follows:\\
- \textbf{point \& normal $(\pmb{p},\vec{n})$}: randomly sample $500$ points on object's surface (along with the corresponding normal directions); \\

\begin{wrapfigure}{r}{0.47\textwidth}
\vspace{-15px}
\label{fig:datagen}
\vspace{-1px}
\begin{minipage}{.47\textwidth}
    \includegraphics[width=1.0\linewidth]{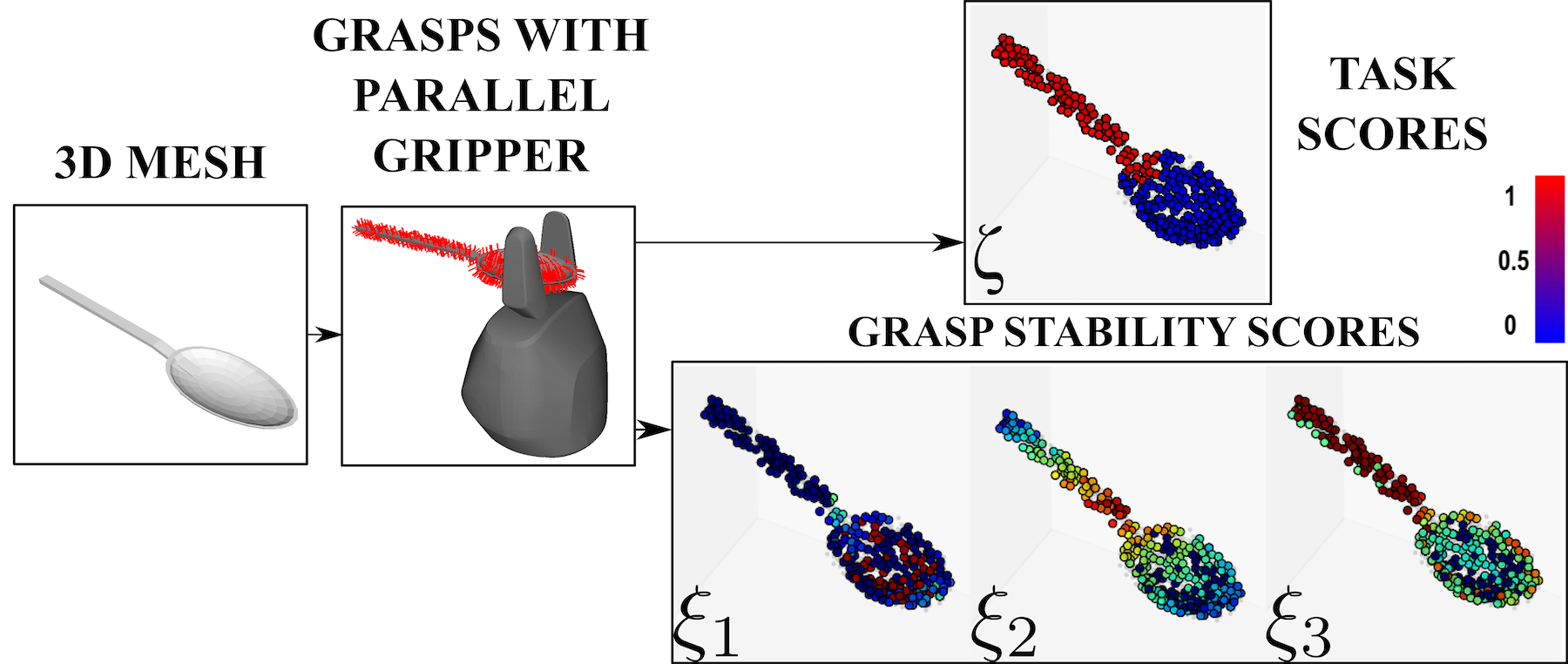}
    \caption{\small{Data generation for \emph{support} on spoon (with $\psi\!=\!0$, $d\!=\!0$).
    For task scores, red area denotes points suitable for grasping.}}
\end{minipage}
\vspace{-20px}
\end{wrapfigure}
\vspace{-16px}
- \textbf{gripper roll $\psi$}: try angles $\psi \in \{0, \pi/2 , \pi/4 \}$; \\
- \textbf{offset $d$}: try offsets $d\!\in\!\{0, 2\text{cm}, 4\text{cm}\}$.
\\
Grasps are simulated by approaching a point on an object along the normal direction and closing the fingers.
Once the fingers close, we extract scores for three of the grasp stability metrics recently analyzed by~\cite{rubert2017relevance}: \\
$\xi_1$ -- Grasp Isotropy Index~\cite{kim2001optimal}; \\
$\xi_2$ -- Distance from Center of Mass~\cite{chinellato2005visual, ponce1995computing, ponce1997computing, ding2001computation}; \\
$\xi_3$ -- Grasp Finger Posture~\cite{cornella2005fast, liu2000computing, li2002analytical}. \\

\vspace{-8px}
\section{Experiments}
\label{sec:experiments}
\vspace{-5px}
\subsection{Performance of BO Variants on Synthetic Benchmarks}
\vspace{-5px}
\label{sec:bo_analytic_tests}

\begin{figure}[t]
\centering
\begin{subfigure}[t]{0.32\textwidth}
\centering
\includegraphics[width=0.8\textwidth]{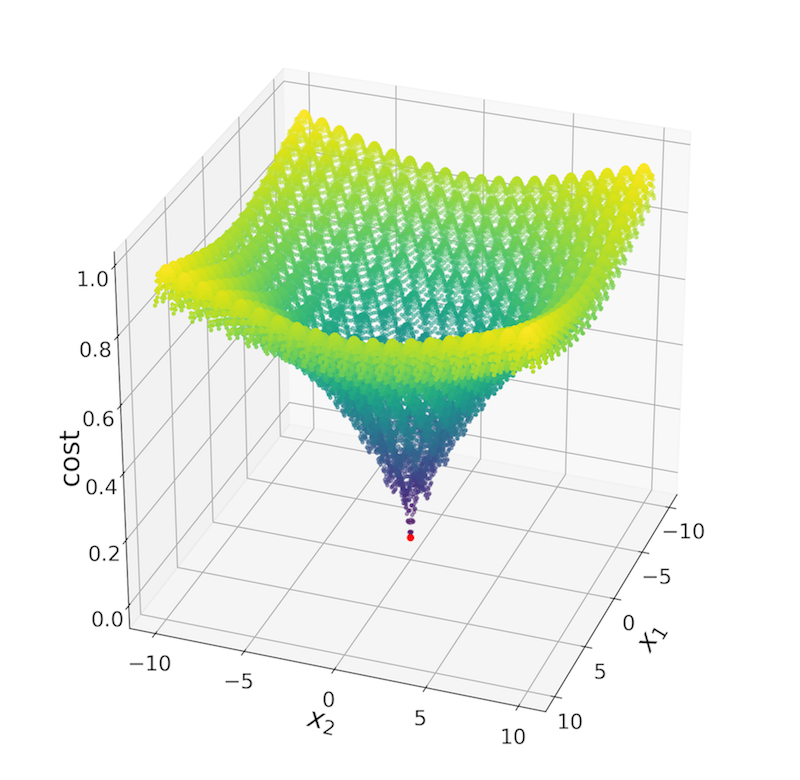}
\caption{\small{Ackley function with \\ challenging local minima.}}
\label{fig:ackley2d}
\end{subfigure}
\hspace{3px}
\begin{subfigure}[t]{0.32\textwidth}
\centering
\includegraphics[width=0.8\textwidth]{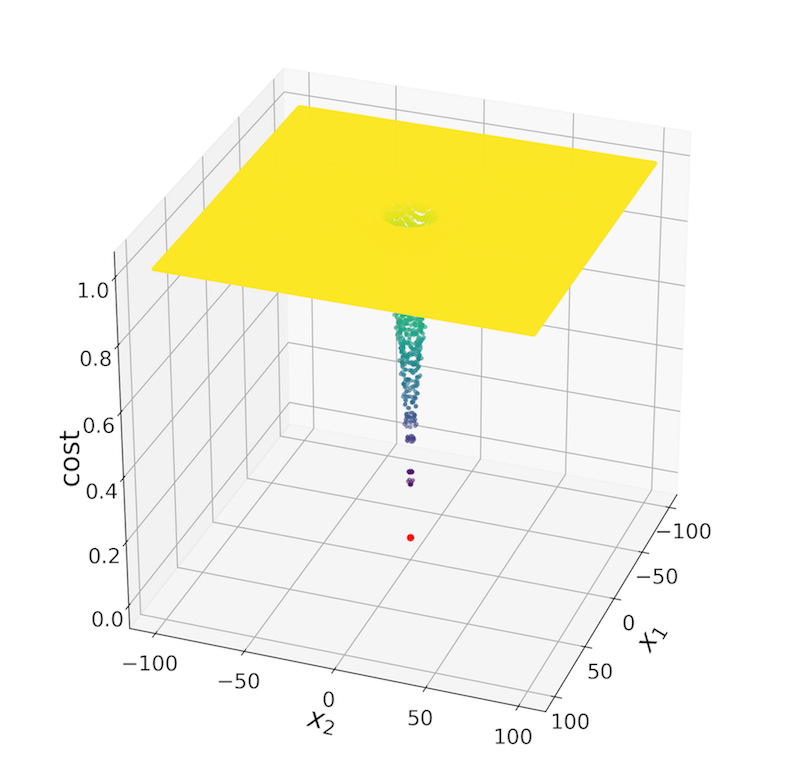}
\caption{\small{Ackley on a larger domain: \\ only 1\% of the space is $<0.9$.}}
\label{fig:ackley2dfar}
\end{subfigure}
\hspace{3px}
\begin{subfigure}[t]{0.32\textwidth}
\centering
\includegraphics[width=0.8\textwidth]{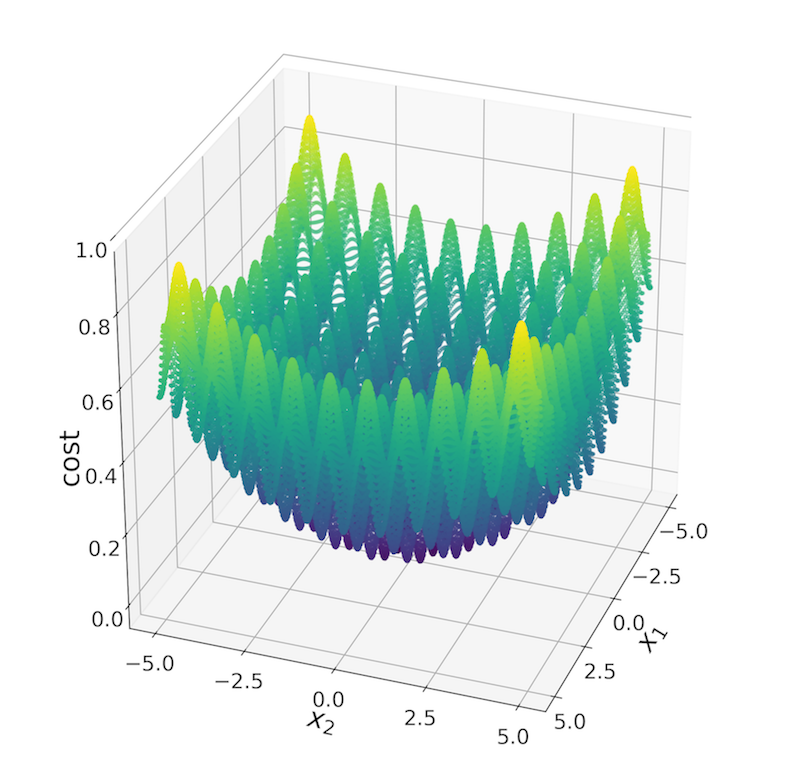}
\caption{\small{Rastrigin function: highly \\ multimodal, deep local minima.}}
\label{fig:rastrigin2d}
\end{subfigure}
\caption{\small{High-dimensional analytic functions for testing optimization algorithms. Ackley and Rastrigin functions present challenges similar to those most frequent in optimization for robotics: (a) shallow local optima, (b) small low cost region \& sharp cost changes, (c) deep local optima difficult to overcome. Figures show functions in 2D, with values (costs) normalized to $[0,1]$. We test on 2- to 10-dimensional versions of these.}}
\label{fig:analytic_functions}
\vspace{-10px}
\end{figure}

To anticipate the challenges of running optimization on a real-world robotics system, we first test the performance of BO variants on synthetic benchmarks. From the commonly used optimization test functions \cite{surjanovic2013virtual} we select 3 settings which exemplify the main challenges that frequently occur in robotics: numerous shallow local optima, small low-cost region and sharp cost function drops/rises, deep local optima difficult to overcome when searching for global optimum. Figure~\ref{fig:analytic_functions} visualizes the settings we consider. Figure~\ref{fig:ackley2d} shows the Ackley function with numerous shallow local minima: \\
$f_{AC}(x) = $\scalebox{0.75}[1.0]{\( - \)}$a \cdot \exp\big($\scalebox{0.75}[1.0]{\( - \)}$b \sqrt{\tfrac{1}{d} \sum_{i\text{=}1}^d x_i^2}\big) - \exp\big(\tfrac{1}{d} \sum_{i\text{=}1}^d\cos(c x_i)\big)+a+\exp(1); \ \ a\!=\!20, b\!=\!0.2, c\!=\!2\pi, x \in \mathbb{R}^d$ \\
For a further challenge we consider this function on a much larger domain $[-100, 100]$. This is similar to considering Easom function, which is commonly used to test robustness to steep ridges/drops in the search space. Figure~\ref{fig:ackley2dfar} shows the Ackley function on this larger domain.

\begin{figure}[t]
\centering
\begin{subfigure}[t]{0.23\textwidth}
\centering
\includegraphics[width=1.0\textwidth]{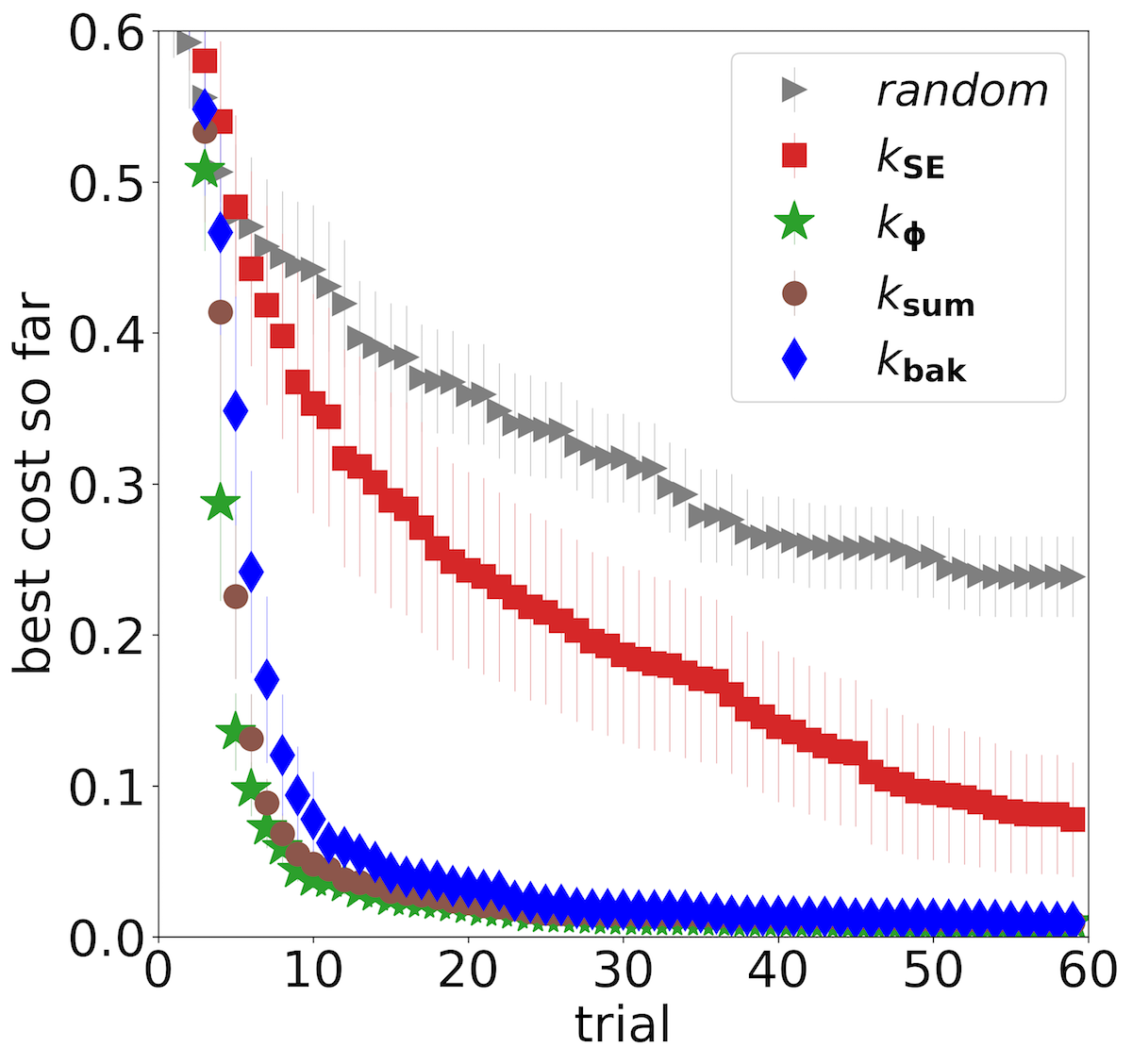}
\caption{\scriptsize{Ackley 2D on $[-10,10]$.}}
\label{fig:ackley2d_runs}
\end{subfigure}
\hspace{3px}
\begin{subfigure}[t]{0.23\textwidth}
\centering
\includegraphics[width=1.0\textwidth]{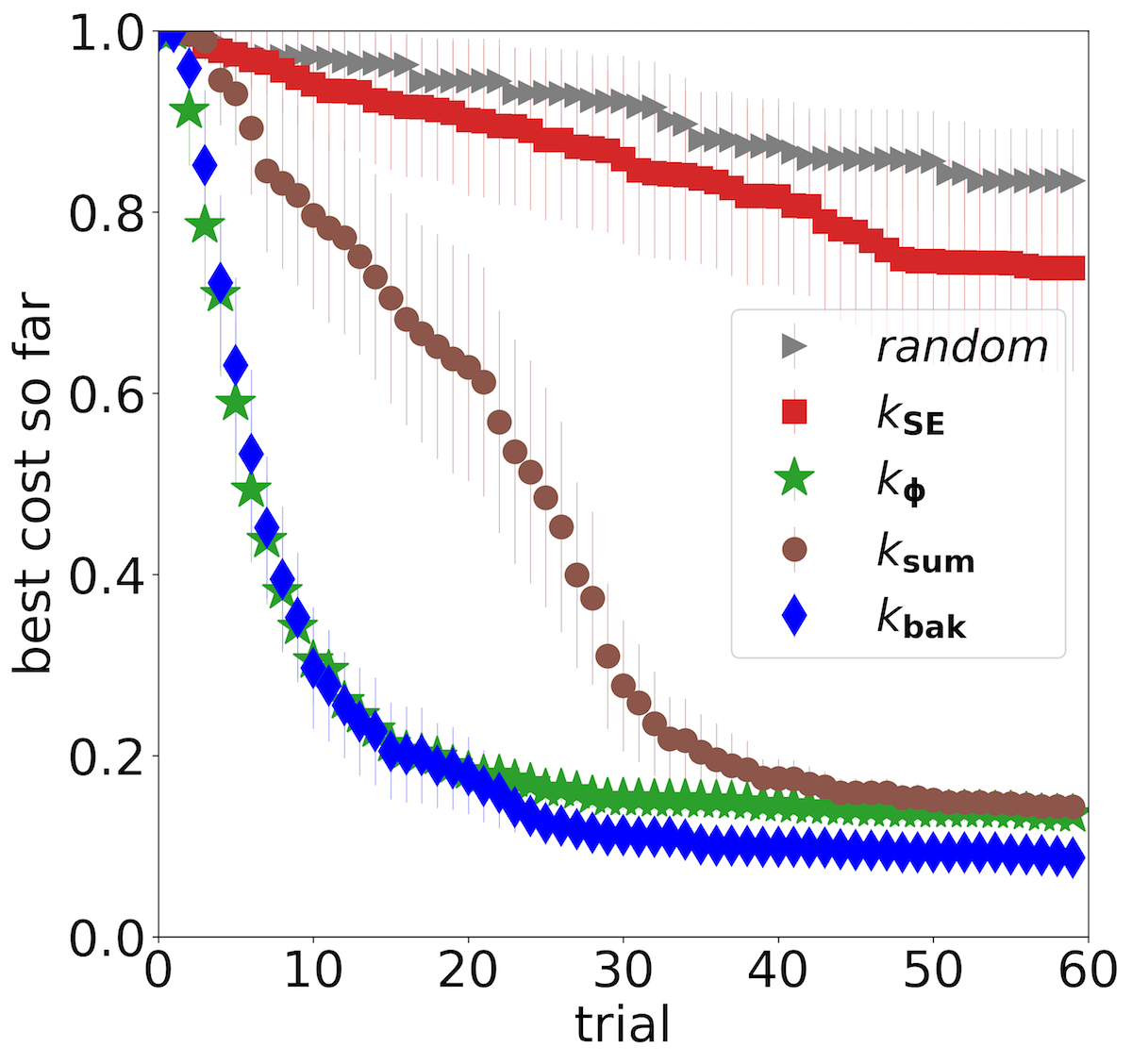}
\caption{\scriptsize{Ackley 2D on $[-100,100]$.}}
\label{fig:ackley2dfar_runs}
\end{subfigure}
\hspace{3px}
\begin{subfigure}[t]{0.23\textwidth}
\centering
\includegraphics[width=1.0\textwidth]{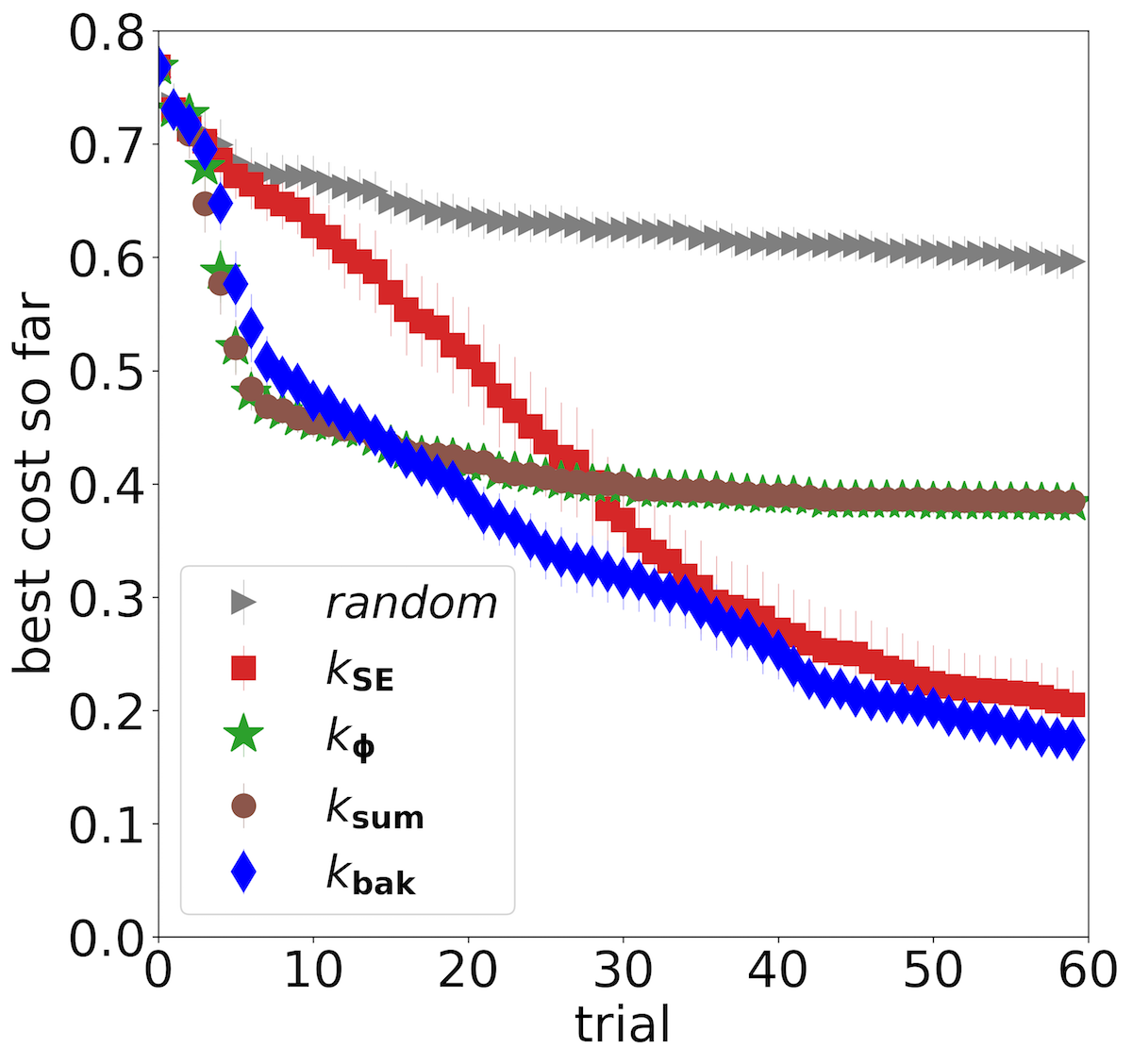}
\caption{\scriptsize{Ackley 10D on $[-10,10]$.}}
\label{fig:ackley2dfar_runs}
\end{subfigure}
\begin{subfigure}[t]{0.23\textwidth}
\includegraphics[width=1.0\textwidth]{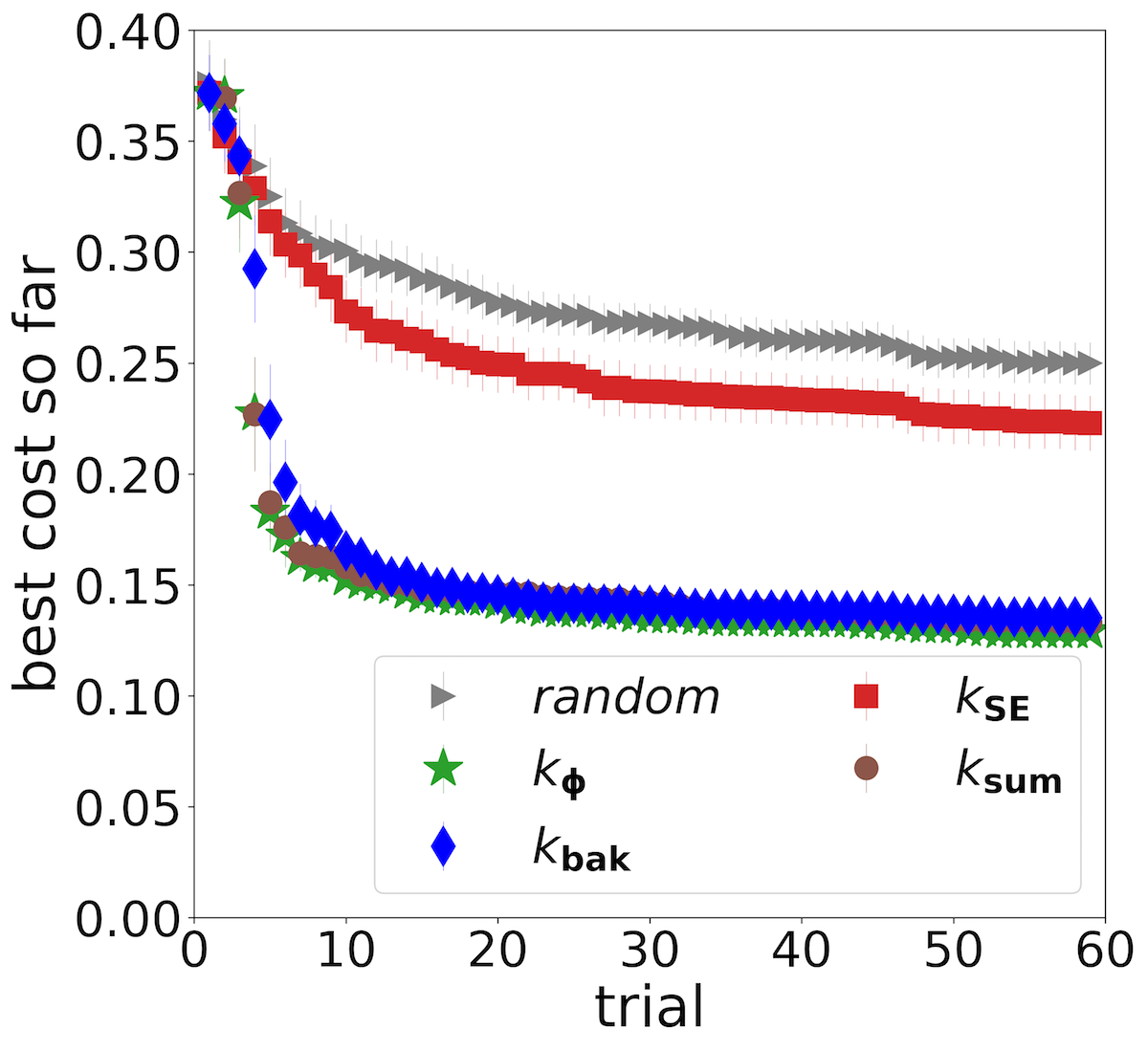}
\caption{\scriptsize{Rastrigin 10D on $[-5,5]$.}}
\label{fig:rastrigin_runs}
\end{subfigure}
\caption{\small{BO on analytic functions. Plots of mean over 40 runs for each kernel type, 95\% CIs. Comparison with random search (`random' in the legend) shows relative difficulty of each setting.}}
\vspace{-15px}
\label{fig:ackley_runs}
\end{figure}

We are interested in emulating a setting where simulation could provide coarse guidance, but also could be limiting when searching for the global optimum. For the synthetic setting we pick $\phi$ with components that capture information from two first addends of the Ackley function:
$[\phi(x)]_1 = $\scalebox{0.75}[1.0]{\( - \)}$b \sqrt{\tfrac{1}{d} \sum_{i\text{=}1}^d x_i^2}; \ [\phi(x)]_2 = \tfrac{1}{d} \sum_{i\text{=}1}^d\cos(c x_i)$.
This gives a kind of `collapsing' of the relevant features, akin to simulation that aims to capture most salient features needed to approximate real-world output. Figure~\ref{fig:ackley_runs} shows comparisons of BO variants on 2D and 10D versions of the Ackley function. In 2D, using $\phi$ gives a significant advantage to all the informed versions of BO (with $k_{\phi}, k_{sum}$ and $k_{bak}$ kernels). When the low-cost region is small, the gains are even more striking. In this case the performance of uninformed BO with SE kernel degrades to random search, while the informed versions get close to the optimum in less than 40 trials. However, the hint of limitations induced by using $\phi$ is already visible: in the case of $k_{\phi}$ and $k_{sum}$ the improvement stagnates after 50 trials. This stagnation is even more striking when 10-dimensional version of the Ackley function is optimized. There, even the uninformed BO with SE kernel improves over $k_{\phi}$ and $k_{sum}$ after 30 trials. In contrast, $k_{bak}$ is able to `recover' and outperform both informed and uninformed kernels.

To test robustness to deep local minima we use Rastrigin function:
$f_{RA}(x) = \sum_i^{d} x_i^2 - \sum_i^{d} a\cos(c\pi x_i) +  a \cdot d; \ a\!=\!10, c\!=\!2, x\!\in\! \mathbb{R}^d$.
Figure~\ref{fig:rastrigin2d} shows a 2D version of the function. As before, $\phi$ summarizes the first two addends:
$[\phi(x)]_1 = \sum_i^{d} x_i^2; \ [\phi(x)]_2 = \sum_i^{d} a\cos(c\pi x_i)$.
Figure~\ref{fig:rastrigin_runs} shows results for the BO variants in 10D. In this case, all informed kernels improve over uninformed SE kernel.
Overall, the results on test benchmarks appear promising. Therefore we proceed with experiments on hardware, since we believe this step is crucial for validating applicability of our approach to robotics problems.

\subsection{Experimental Setup for Task-oriented Grasping}
\begin{figure}[h]
\begin{minipage}{.52\textwidth}
\centering
\renewcommand{\tabcolsep}{1pt}
    \begin{tabular}{|c|c|c|c|}
    \hline
    \begin{minipage}{.25\textwidth}
    \includegraphics[width=\linewidth, width=18mm,height=11mm]{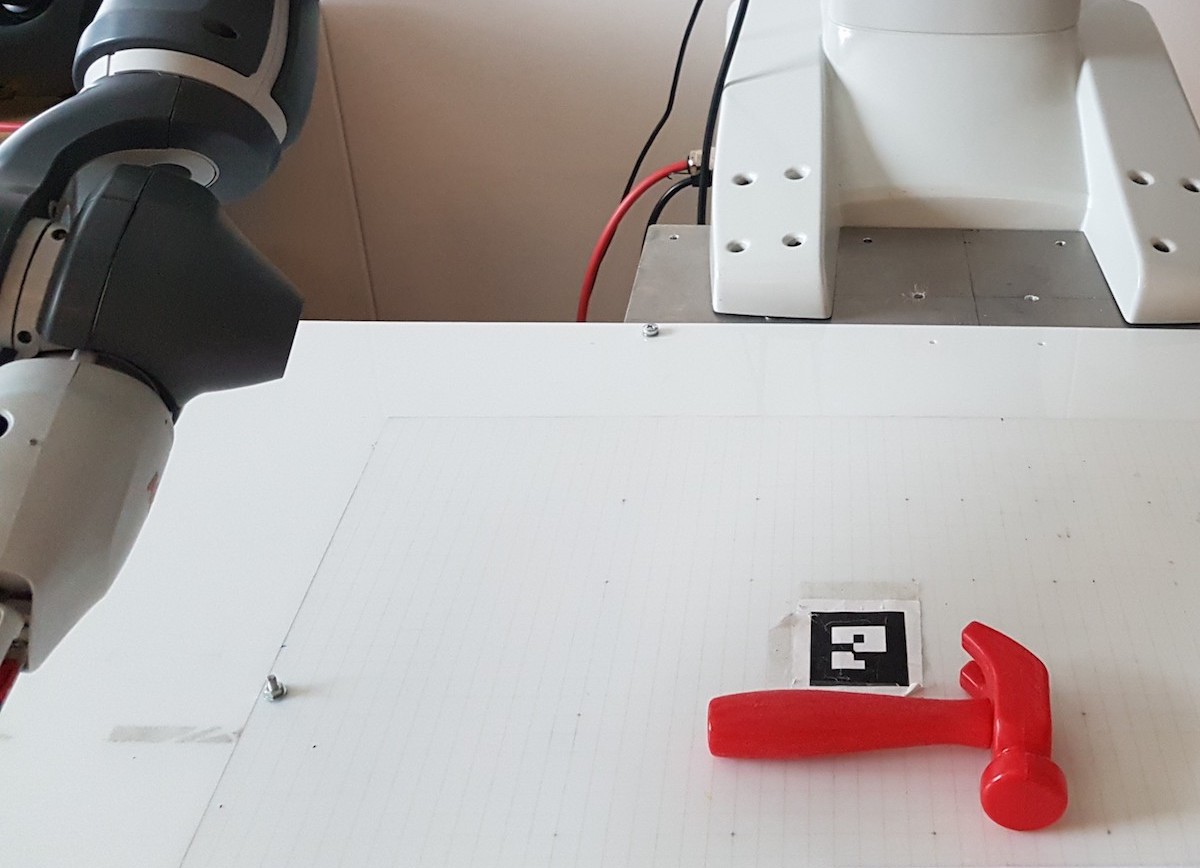}
    \includegraphics[width=\linewidth, width=18mm,height=11mm]{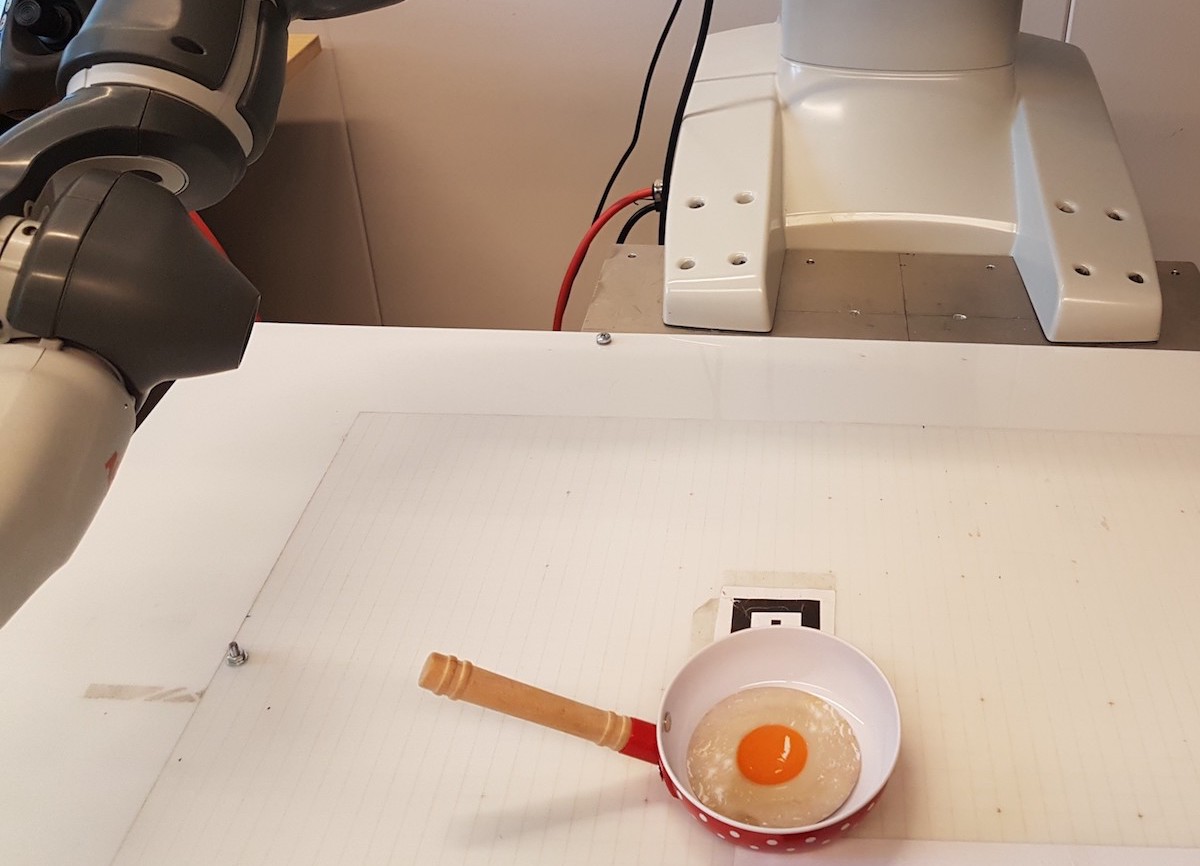}
    \includegraphics[width=\linewidth, width=18mm,height=11mm]{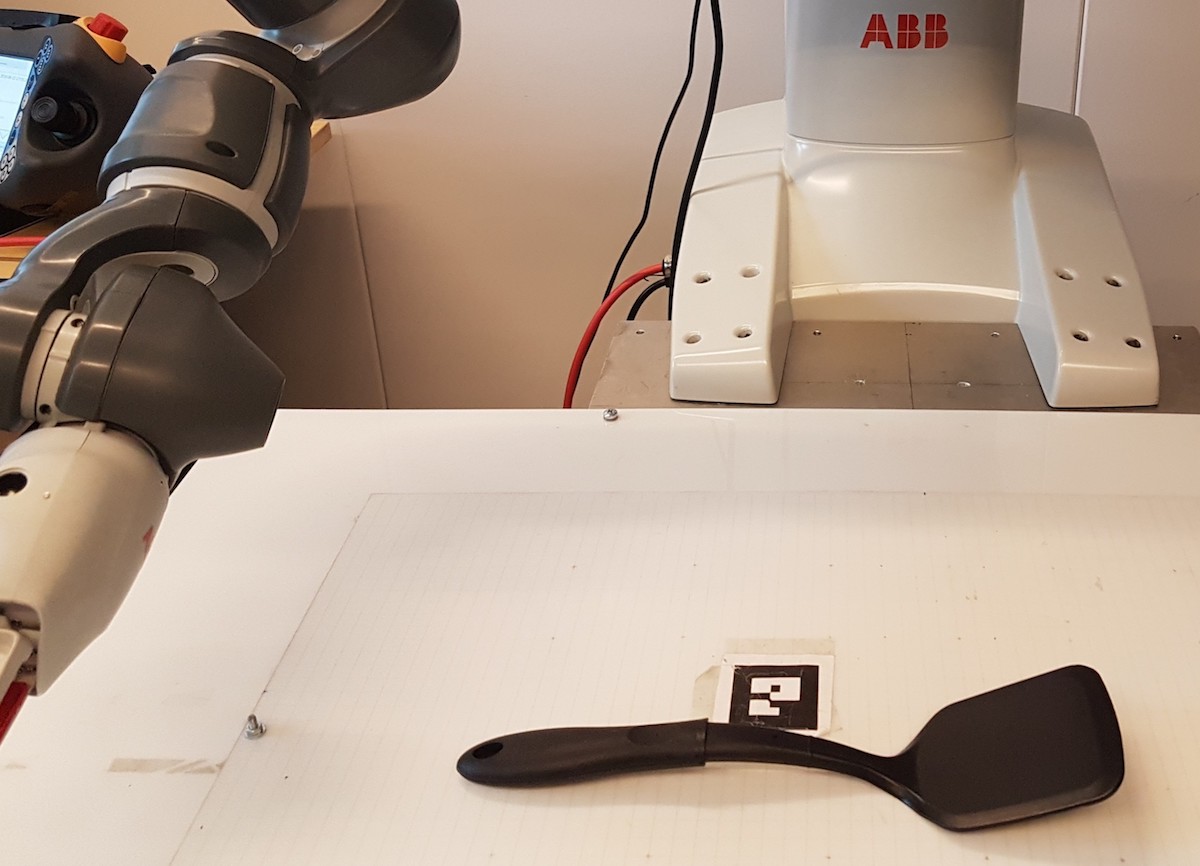}
    \end{minipage}
    &
    \begin{minipage}{.21\textwidth}
    \includegraphics[width=\linewidth, width=15mm]{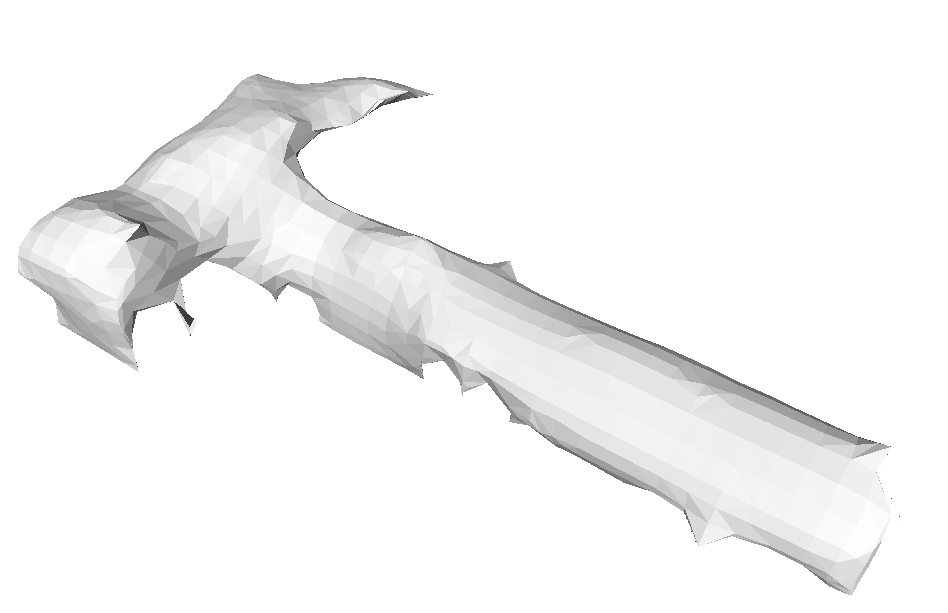}
    \includegraphics[width=\linewidth, width=15mm]{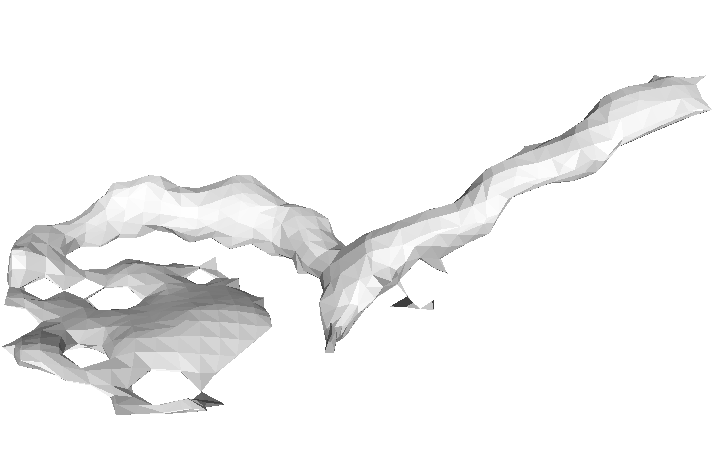}
    \includegraphics[width=\linewidth, width=15mm]{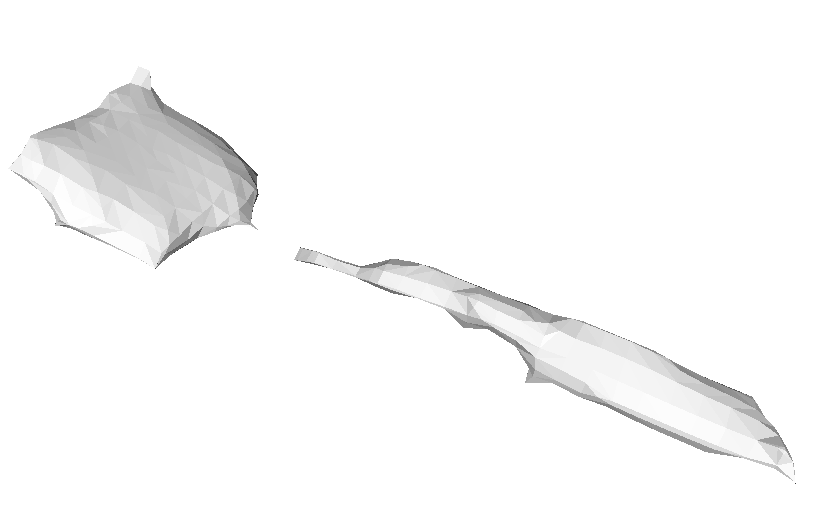}
    \end{minipage}
    &
    \begin{minipage}{.21\textwidth}
    \includegraphics[width=\linewidth, width=15mm]{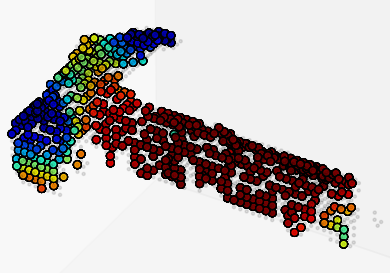}
    \includegraphics[width=\linewidth, width=15mm]{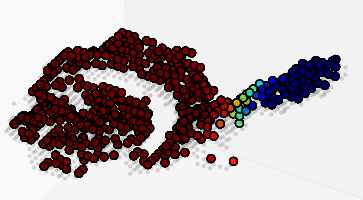}
    \includegraphics[width=\linewidth, width=15mm]{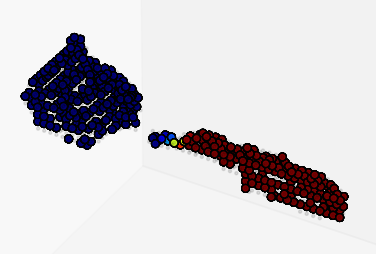}
    \end{minipage}
    &
    \begin{minipage}{.21\textwidth}
    \includegraphics[width=\linewidth, width=15mm]{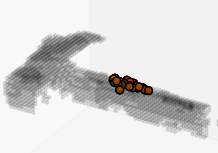}
    \includegraphics[width=\linewidth, width=15mm]{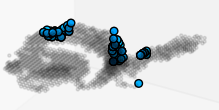}
    \includegraphics[width=\linewidth, width=15mm]{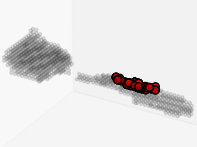}
    \end{minipage}\\
    \hline
    \end{tabular}
    \label{fig:ymtg}
    \caption{\small{Three objects from our everyday objects dataset (EOD) and visualization of task \& stability scores. From left to right: object in the workspace, partial mesh, $\zeta$ task scores, $\xi_2$ stability metric for top $100$ grasps with $\zeta\!>\!0.5$.}}
\end{minipage}
\hspace{5px}
\begin{minipage}{.42\textwidth}
\renewcommand{\tabcolsep}{1pt}
\renewcommand{\tabrowsep}{1pt}
\vspace{-8px}
    \scriptsize
    \begin{tabular}{ | c | c | c | }
    \hline
    \begin{minipage}{.30\textwidth}
      Black pan
      \includegraphics[width=\linewidth, height=10mm]{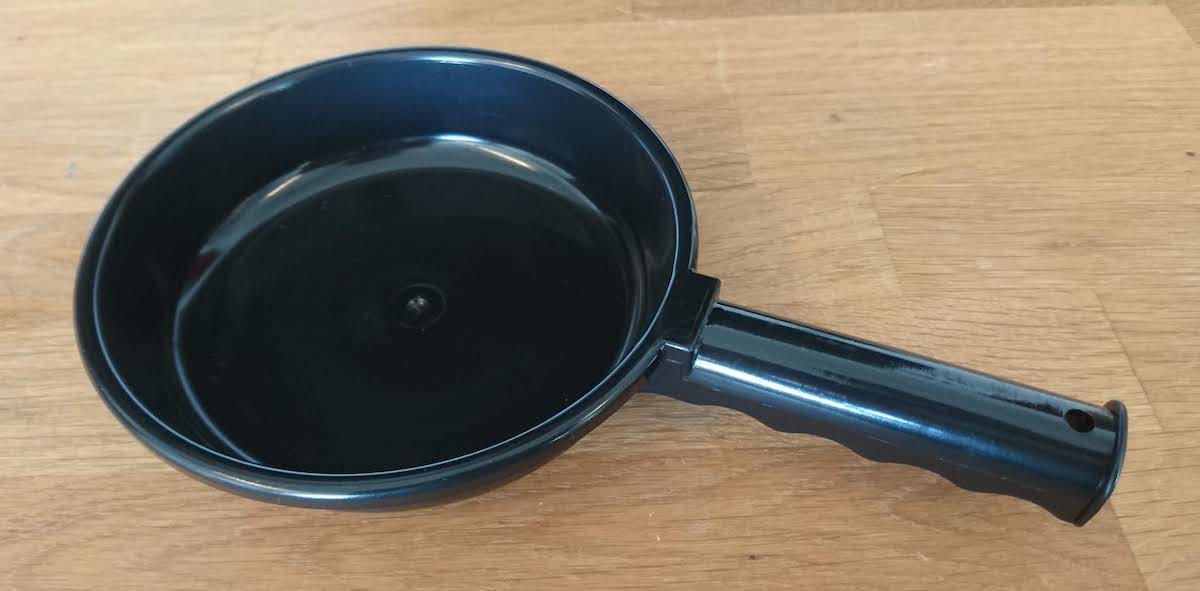}
      Red pan
      \includegraphics[width=\linewidth, height=10mm]{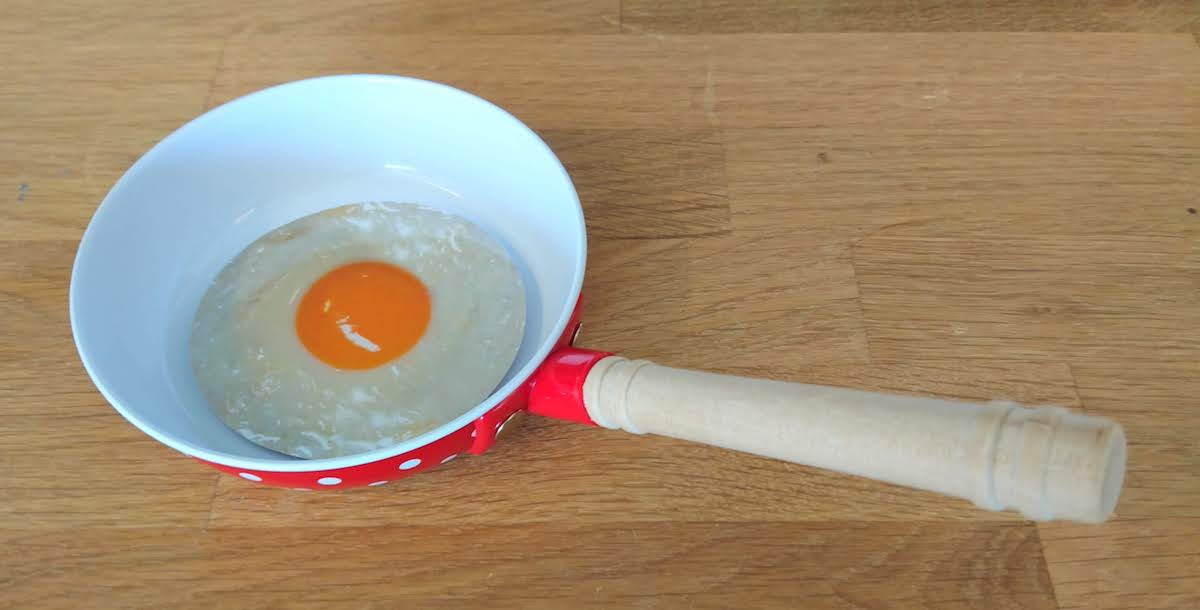}
      Knife
      \includegraphics[width=\linewidth, width=17mm,height=10mm]{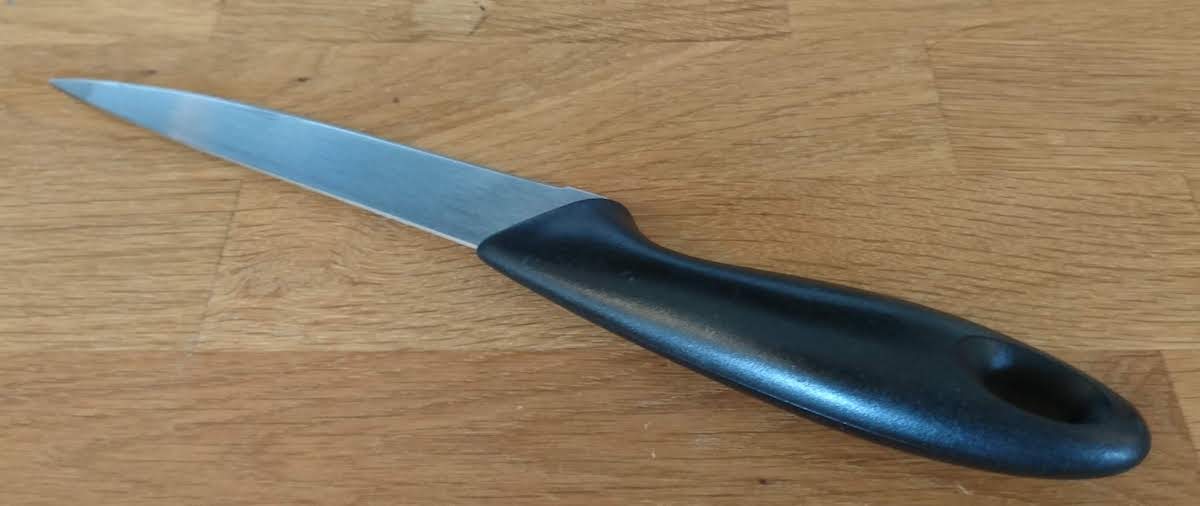}
    \end{minipage}
    &
    \begin{minipage}{.40\textwidth}
      Red screwdriver
      \includegraphics[width=\linewidth, height=11mm]{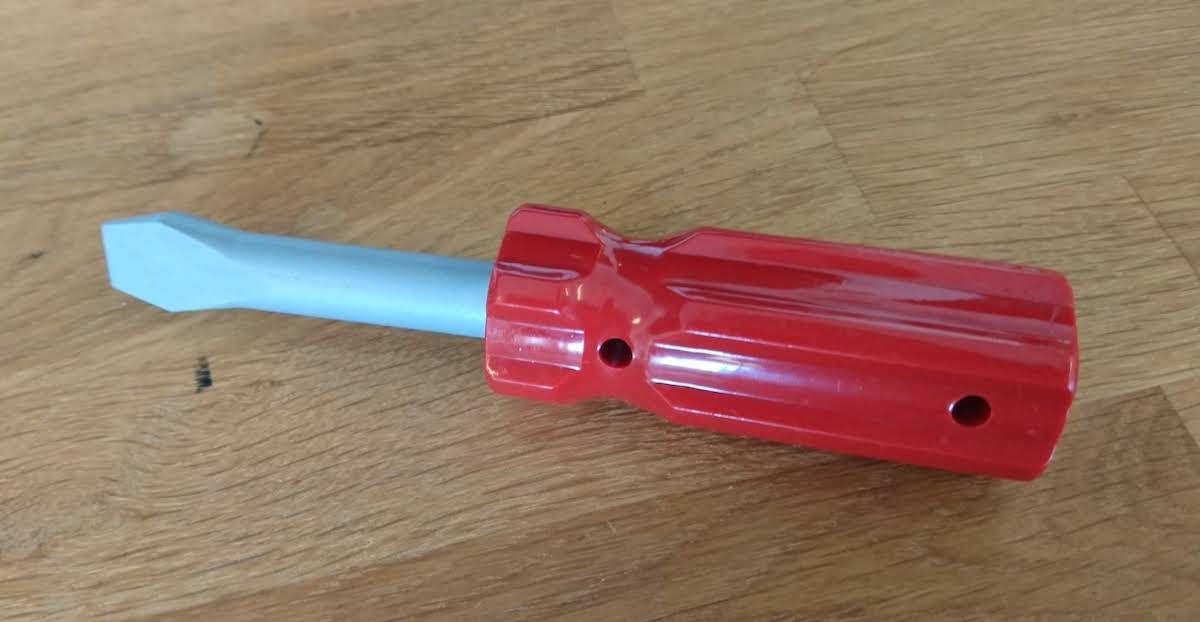}
      Black screwdriver
      \includegraphics[width=\linewidth, height=11mm]{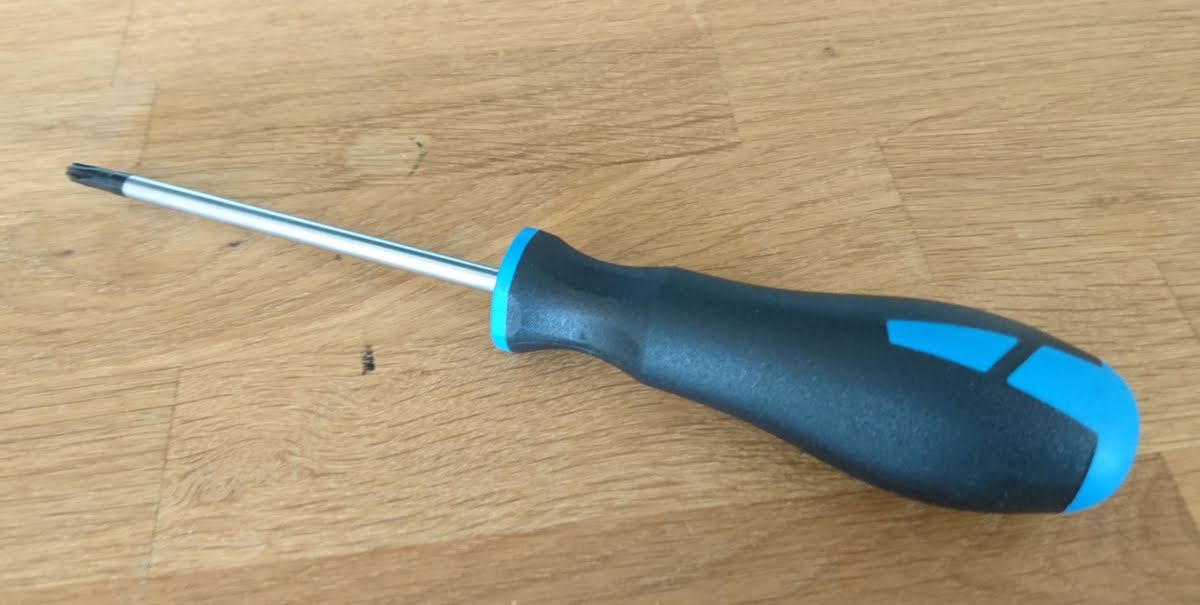}
      Spatula
      \includegraphics[width=\linewidth, width=23.5mm]{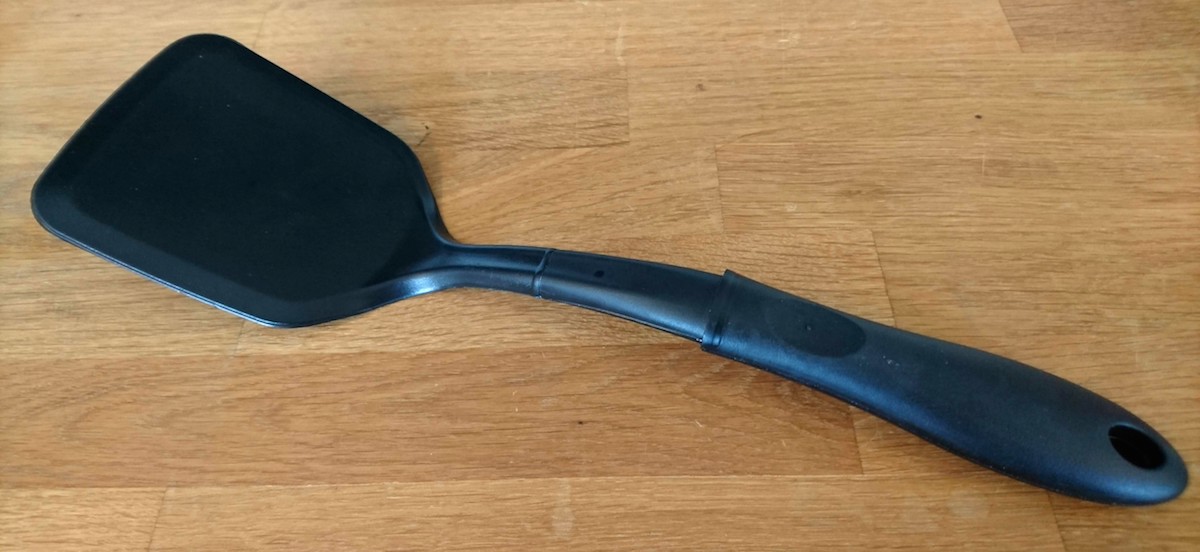}
    \end{minipage}
    &
    \begin{minipage}{.33\textwidth}
      Red hammer
      \includegraphics[width=\linewidth, height=10mm]{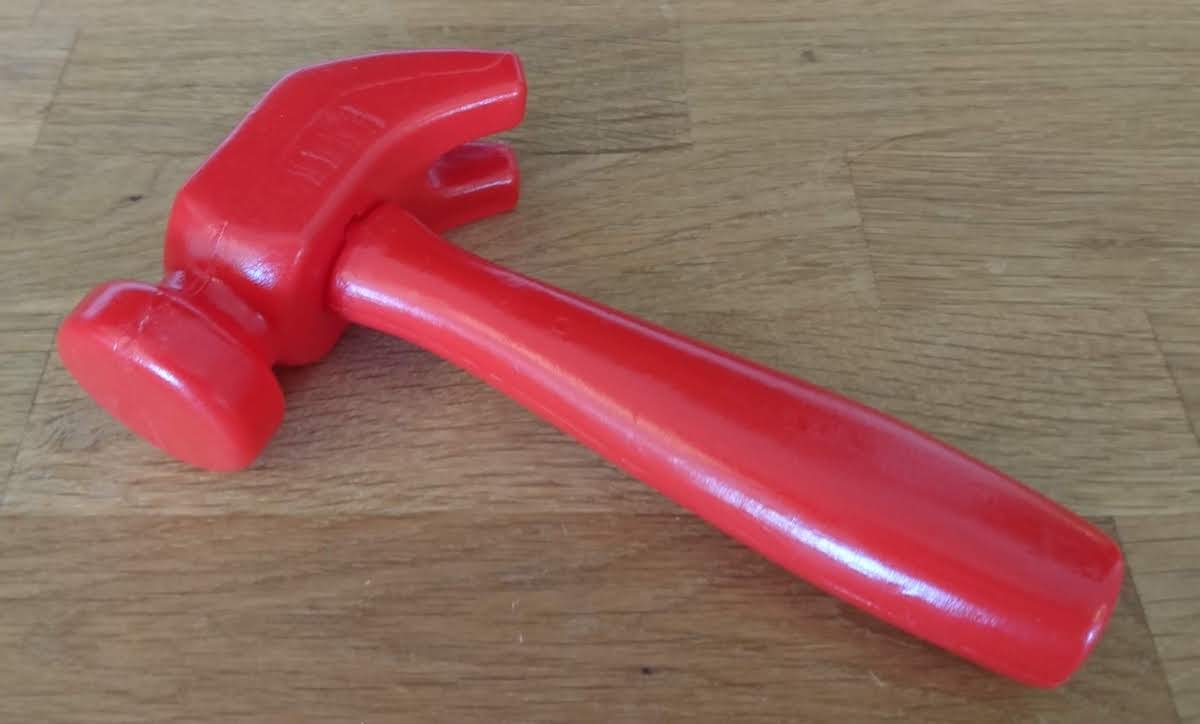}
      Black hammer
      \includegraphics[width=\linewidth, height=10mm]{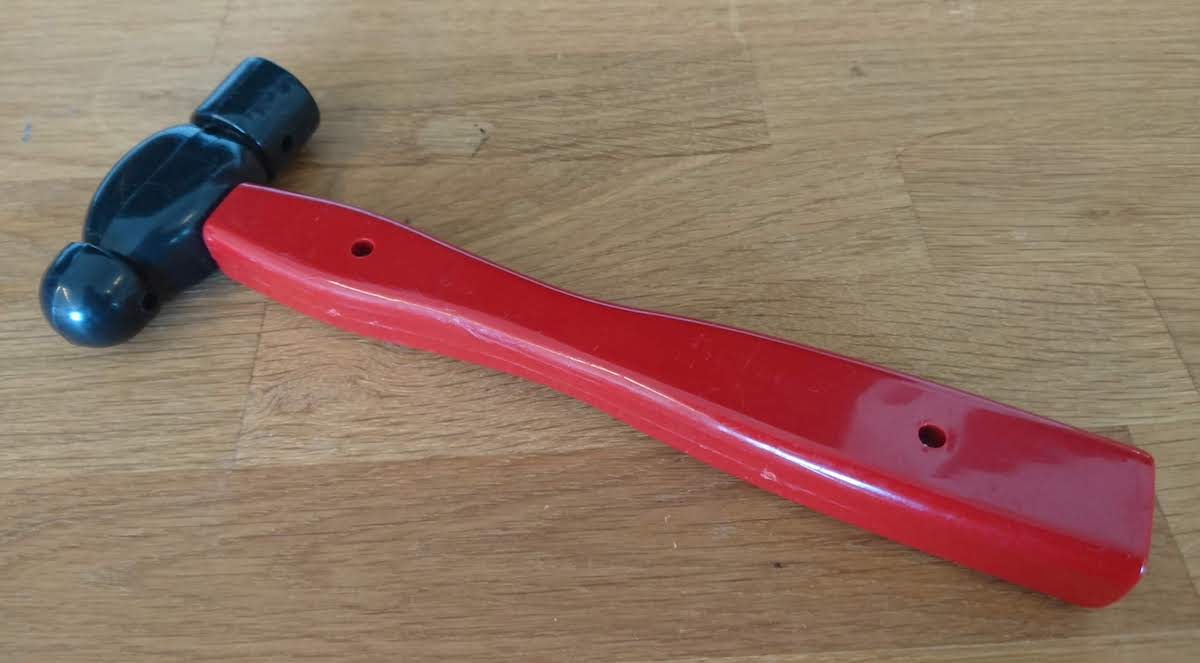}
      Pen $\quad \quad \ \ \ $ Mug
      \hspace{2px}
      \includegraphics[width=\linewidth, width=6.5mm]{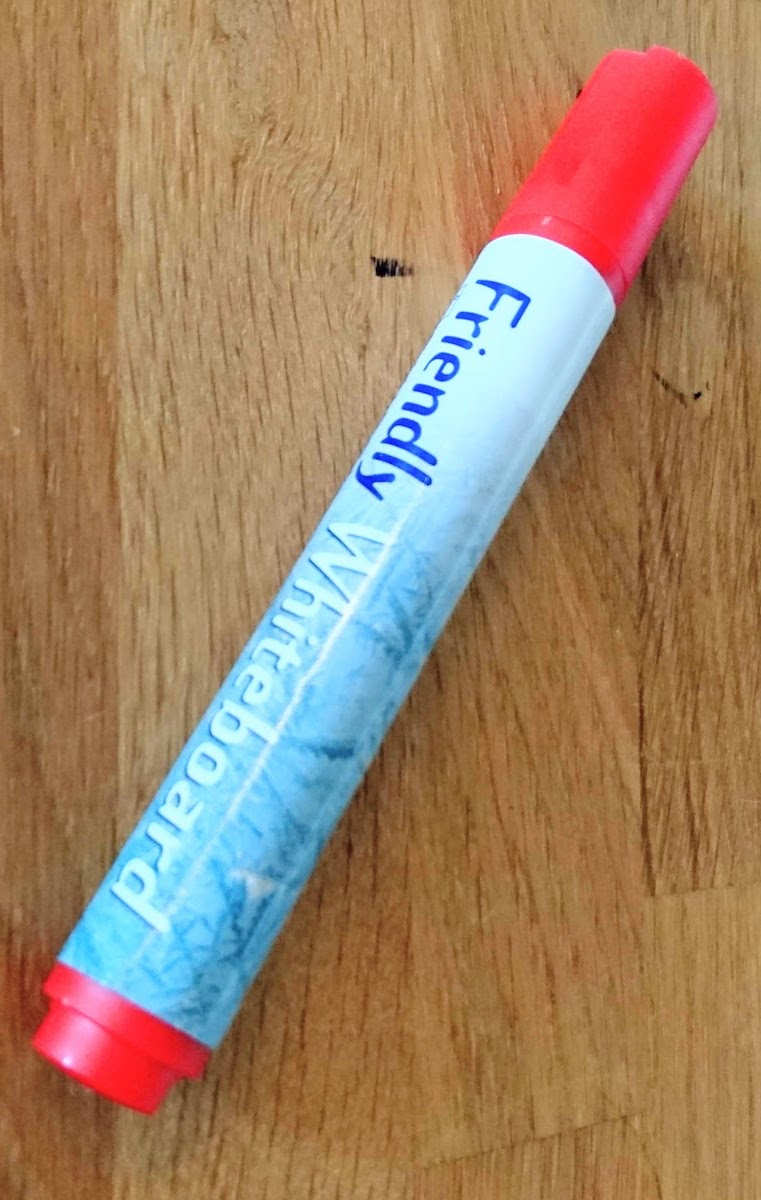}
      \hspace{1px}
      \includegraphics[width=\linewidth, height=10mm,width=11mm]{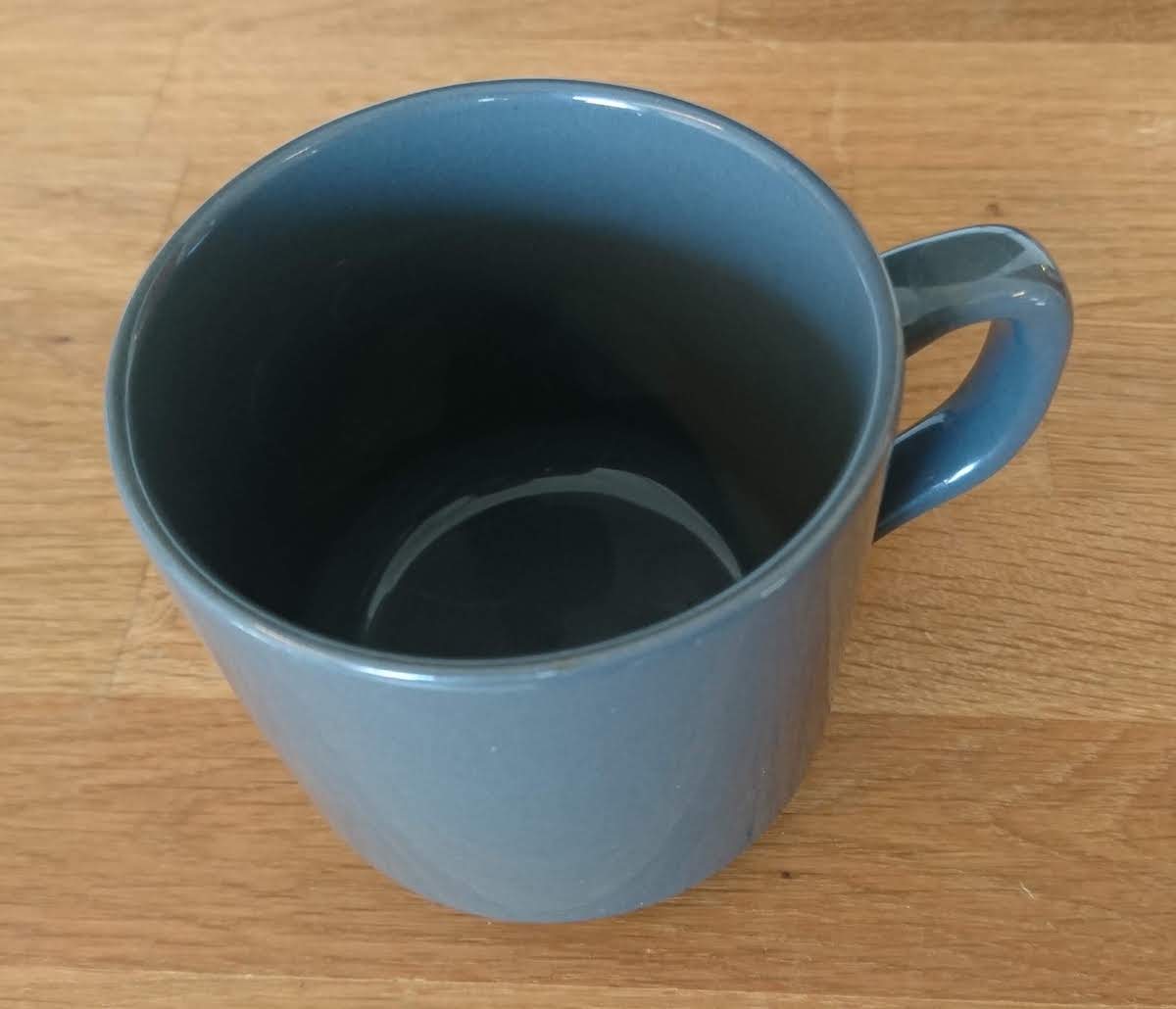}
    \end{minipage}
    \\ \hline
  \end{tabular}
  \vspace{-2px}
  \caption{\small{Ten objects from our EOD.}}
  \label{fig:eod}
\end{minipage}
\end{figure}

\vspace{-5px}
\label{sub:des}
To evaluate performance on a real-world setting we construct Everyday Objects Dataset (EOD), then do task-oriented grasping with an ABB Yumi robot. The objects (shown in Figure~\ref{fig:eod}) are from seven categories and can be used for six different tasks. Our selection of objects is constrained by limitations of the robot: maximum payload of $500$g (lower in practice), rigid plastic parallel gripper, no force-torque or tactile sensing.
Once an object is placed on the table, we use Microsoft Kinect to get a dense point cloud of the scene and segment out the object. We then generate a mesh representation from the partial point cloud and attach a coordinate frame to the object\footnote{Object coordinate frame is positioned in the center of the mesh bounding box and has the same orientation in the world coordinate frame as the april tag which we use to segment the object, for details see \cite{varley2017shape}.}. Then we compute grasp points, discarding those in collision with the table. The object is voxelized and together with grasp and task representation fed through the CNN to get stability and task score estimates. 

We execute grasps on the robot and report the results using the following scores:\\
\textbf{[0.0-0.99] No plan}: planning failed or returned partial plan; \\
\textbf{[1.0] No lift}: failed to grasp the object; \\
\textbf{[2.0] Lift failed}: the robot grasped the object but failed to lift it up (the object slipped); \\
\textbf{[3.0] Wrong grasp}: grasped the wrong part of the object (unsuitable for the task); \\
\textbf{[4.0] Success}: task-appropriate grasp was successful. \\
After each execution, operator re-positions the object in the workspace if needed and starts next trial. 

\begin{wraptable}{r}{.41\textwidth}
\vspace{-20px}
\centering
\scalebox{0.8}{
    \begin{tabular}{c c c c}
    \\ \toprule
    Task	&  \textbf{maxF1} & \textbf{MAP}\\
    \midrule
    handover	& 0.892 & 0.930\\
    screw    & 0.988 & 0.990\\
    cut    & 0.764 & 0.877\\
    pour	& 0.874 & 0.963\\
    support	& 0.952 & 0.964\\
    pound	& 0.940 & 0.948\\ 
    \midrule
    \textbf{overall} & 0.909 & 0.943  \\
    \bottomrule
    \end{tabular}
} 
\scalebox{0.8}{
    \begin{tabular}{c c c c}
    \\ \toprule
    Stability scores & $\xi_1$ & $\xi_2$ & $\xi_3$\\
    \midrule
    RMSE	& 0.184 & 0.173 & 0.206\\
    Precision@$100$ & 0.651 & 0.902 & 0.688\\ 
    \bottomrule
    \end{tabular}
} 
\caption{\small{CNN evaluation on synthetic testset.}}
\label{tab_synthetic_eval}
\vspace{-20px}
\end{wraptable}
Before starting experiments on hardware, for an initial test of our CNN we evaluated its performance on a synthetic test set. For task suitability: we report task-specific and overall MaxF1 and MAP scores in Table~\ref{tab_synthetic_eval} (top). Overall, the network learns to recognize parts of objects that are suitable for grasping given a task. The most challenging task is \emph{cut}, since blade and handle parts of a knife can have very similar appearance.
For grasp stability: in Table~\ref{tab_synthetic_eval} (bottom) we report RMSE for each of the stability metrics.
Since we are interested in using the predicted scores for task-oriented grasping, we also report precision at top $100$ for task-appropriate points ($\zeta > 0.5$).

\subsection{Hardware Experiments for Top-k Grasps}
\label{sub:topk}

\begin{wraptable}{r}{.29\textwidth}
{
\vspace{-23px}
\scalebox{0.8}{
\renewcommand{\tabcolsep}{1pt}
\renewcommand{\tabrowsep}{1pt}
\begin{tabular}{cc}
 \\ \toprule
 \multicolumn{1}{c}{Task} & \multicolumn{1}{c}{Scores (for each trial)}
 \\ \midrule
  \multicolumn{2}{c}{Spatula}  \\
  support & 4 4 4 4 4 2 4 4 4 1 \\
  \midrule
  \multicolumn{2}{c}{White pan}  \\
  handover & 4 4 4 4 4 4 3 3 4 3\\
  support & 4 4 4 4 4 1 4 1 4 4\\
  \midrule
  \multicolumn{2}{c}{Black pan}  \\
  handover & 1 2 2 1 4 2 4 4 4 4\\
  support & 1 1 1 1 4 4 1 2 1 4\\
  \midrule
  \multicolumn{2}{c}{Red hammer} \\
  handover & 1 2 4 4 2 4 2 4 4 1\\
  pound & 4 4 4 4 4 1 4 1 4 4\\
  \midrule
  \multicolumn{2}{c}{Black hammer} \\
  handover & 4 4 4 4 2 4 4 1 1 2\\
  pound & 4 4 4 2 4 4 1 1 1 4\\
  \midrule
  \multicolumn{2}{c}{Mug} \\
  handover & 2 2 4 2 4 2 2 4 2 2\\
  pour & 4 4 2 1 4 3 4 4 4 4\\
  \midrule
  \multicolumn{2}{c}{Red screwdriver} \\
  screw & 4 4 4 1 4 4 2 4 4 2\\
  \midrule
  \multicolumn{2}{c}{Black screwdriver} \\
  screw & 2 2 4 2 1 4 2 2 2 2\\
  \midrule
  \multicolumn{2}{c}{Knife} \\
  cut & 1 2 4 2 4 2 2 4 2 4\\
  \midrule
  \multicolumn{2}{c}{Pen} \\
  handover & 4 2 4 2 4 2 4 4 2 2 
  \\ \bottomrule
\end{tabular}
} 
\caption{Top-10 grasps on EOD.}
\label{tbl:eod_topk}
\vspace{-17px}
}
\end{wraptable}

To test the performance of our CNN in a real-world setting, we did task-oriented grasping for objects in our EOD.
We generated $4500$ grasps (as in Section~\ref{sub:dgs}), then removed those in collision with the table. For each object-task pair, 10 grasps with highest stability according to score $\xi_2$ (and task scores $\zeta\!>\!0.5$) were executed on the robot. We did not perform \emph{handover} for screwdrivers and knife, because their blades were not adequately visible to our camera. Furthermore, although some high-scoring points were found on the spatula for \emph{handover}, they were out of reach for the robot. For all other object-task pairs multiple successful task-oriented grasps were found. Table~\ref{tbl:eod_topk} reports scores for each trial (scoring described in Section~\ref{sub:des}).

Our CNN mostly outputs high scores for grasp points that are on suitable parts of the object, both from task and stability perspectives.
Of course not all of these grasps succeed in practice, since the CNN is trained using simulations that do not model friction or mass distribution. In reality, the most stable grasps were on spatula for \emph{support}, white pan for both tasks and red hammer for task \emph{pound}.
The most unstable grasps where on mug \emph{handover}, likely because of smooth surface, high mass, severely degraded partial mesh from vision.

For the above experiments, we reduced maximum gripper standoff from $4$cm to $3$cm. Since the tabletop was not present during training, CNN was likely to learn high scores for grasps with small standoff. These would frequently collide with the table. The alternative grasps with $4$cm standoff were still successful in simulation, but slipped off the object in reality. While we could fix this issue by simply reducing the maximum standoff in this case, such simulation-reality mismatch might be harder to fix in general. This motivates experiments with adaptive search like BO, which we describe next.
\vspace{-2px}
\subsection{Hardware Experiments for Bayesian Optimization}
\label{sub:expbo}
\vspace{-5px}

We ran experiments using the proposed BO approach from Section~\ref{subsec:approach_bo}. $k_{\phi}$ kernel was constructed using the stability scores: $\phi(\pmb{x}) = [\xi_1, \xi_2, \xi_3]$, which were obtained from the trained CNN. The vector of control parameters $\pmb{x}\!=\!(\pmb{p},\vec{n}, \psi, d)$ contained: 3D coordinates for a point on the object, approach direction, gripper roll and offset (as in Section~\ref{sub:TOG}). Choice for $\pmb{p}$ was constrained to the surface of the object, but other control parameters were limited only by a choice of [min,max] values. Points on the object were sampled from the output of the vision system, those with low task scores were filtered out (for a given task) using the trained CNN. Our implementation of BO was based on~\cite{GPMLCode, gardner2014bayesian}. After each BO trial, $f(\pmb{x})$ evaluation (expressing the success of executing grasp with parameters $\pmb{x}$) was given as described in Section~\ref{sub:des}.

\begin{wrapfigure}{r}{0.32\textwidth}
\vspace{-12px}
\centering
\begin{subfigure}[t]{0.15\textwidth}
\centering
\includegraphics[width=1.0\textwidth]{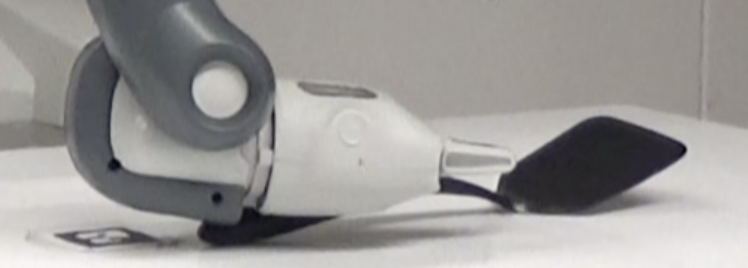}
\includegraphics[width=1.0\textwidth]{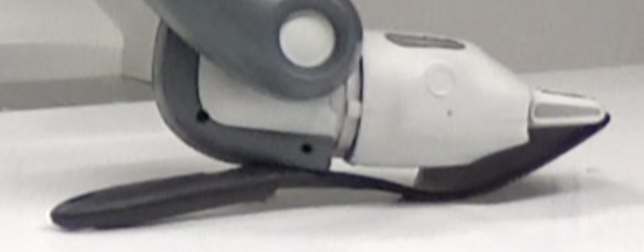}
\includegraphics[width=1.0\textwidth]{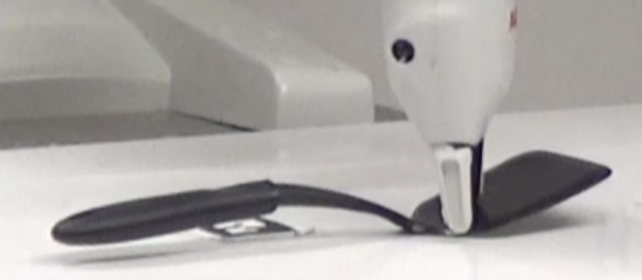}
\label{fig:spatula_p_handover_net}
\end{subfigure}
\hspace{1px}
\begin{subfigure}[t]{0.15\textwidth}
\centering
\vspace{-22px}
\includegraphics[width=1.0\textwidth]{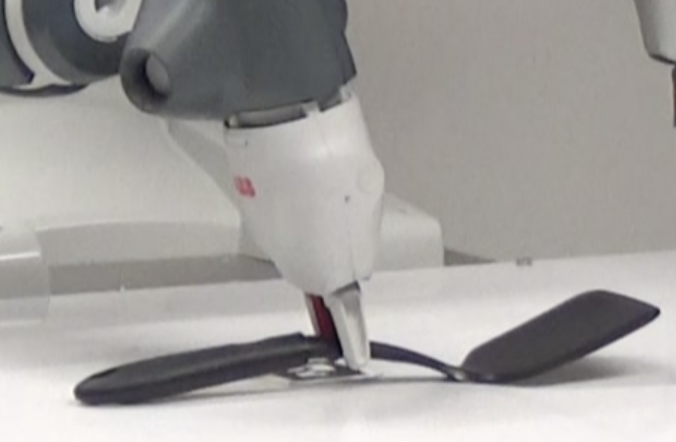}
\includegraphics[width=1.0\textwidth]{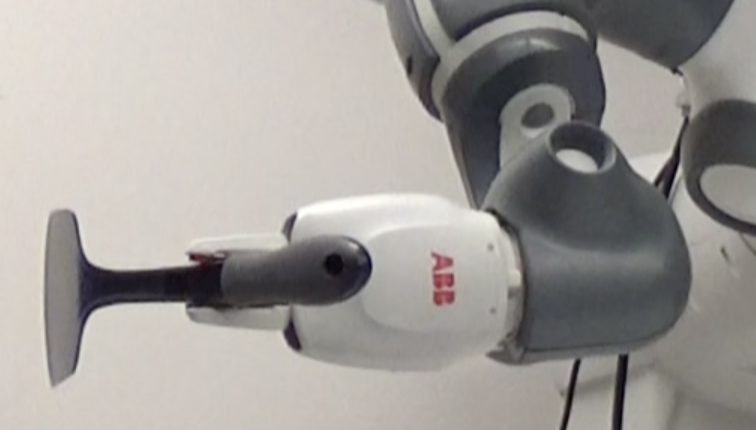}
\label{fig:spatula_p_handover_bo}
\end{subfigure}
\vspace{-12px}
\caption{\small{Left: top choices for handover task from 3 stability metrics; Right: subsequent BO trial.}}
\label{fig:spatula_p_handover}
\vspace{-7px}
\end{wrapfigure}
BO usually starts with a random trial, but in our experiments we instead execute one top choice from each of the 3 stability metrics. For challenging objects this does not yield any promising points. However, from this BO can infer which regions are not promising. 
\emph{Handover} task for spatula provides a clear example of this. Left side of Figure~\ref{fig:spatula_p_handover} shows executing top choice according to each of the 3 stability metrics (after filtering out planning failures). These choices are not successful, moreover executing top 20 choices does not yield successful grasps. Most stable grasps are not reachable due to robot's joint limits or the need to approach too close to the table. We obtain a successful handover on the $4^{th}$ trial (1st candidate from BO), shown in the right part of Figure~\ref{fig:spatula_p_handover}. The grasp is just above the handle part (the handle needs to be clear for handover). It is labeled as acceptable for task completion, but is unlikely to be among top $k$ offline choices. This is because task and stability scores are higher in other parts of the object, but those parts are inaccessible.

\begin{wrapfigure}{r}{0.32\textwidth}
\begin{subfigure}[t]{0.15\textwidth}
\centering
\vspace{-12px}
\includegraphics[width=0.9\textwidth]{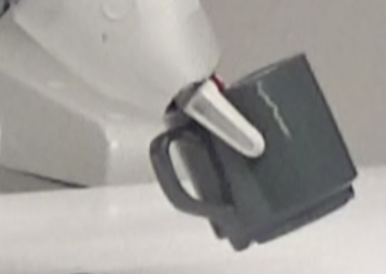}
\includegraphics[width=0.9\textwidth]{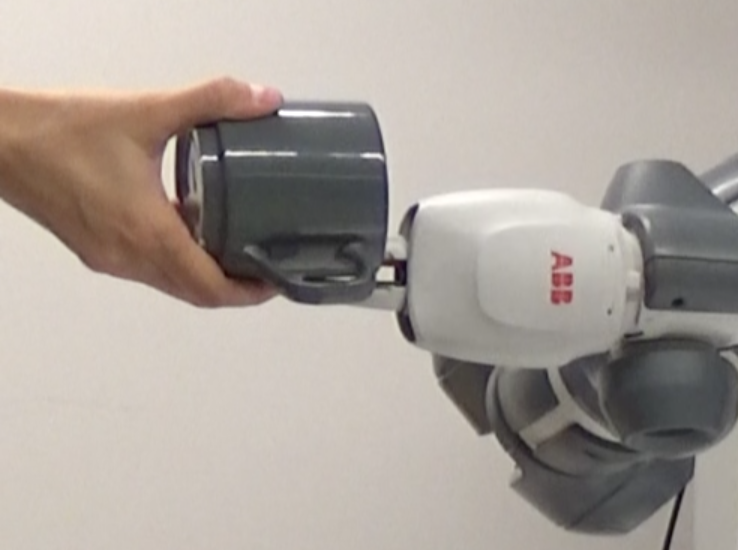}
\label{fig:mug_p_handover_bo}
\end{subfigure}
\hspace{1px}
\begin{subfigure}[t]{0.15\textwidth}
\centering
\vspace{-12px}
\includegraphics[width=1.0\textwidth]{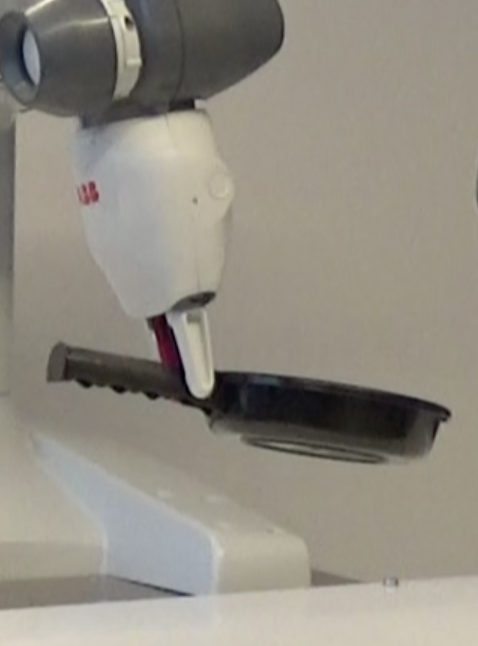}
\label{fig:pan_p_support_bo}
\end{subfigure}
\vspace{-15px}
\caption{\small{Left: A trial from BO for mug handover task. Right: A trial from BO for pan support task.}}
\label{fig:bo_trials_mug_pan}
\vspace{-15px}
\end{wrapfigure}
Left side of Figure~\ref{fig:bo_trials_mug_pan} shows success of BO on the $4^{th}$ trial for mug handover (1st candidate from BO). 
Right side shows successful pan handover on the $5^{th}$ trial (2nd candidate from BO). In contrast, top $k$ grasps computed offline attempted to grasp the middle and outer part of the handle, which for this pan resulted in either slippage or tilting.
The above object-task pairs presented the toughest challenges for the top-10 approach, while BO proposed successful task-oriented grasps in the first few trials.

Overall, we did 15 runs with BO: 6 full runs with 10 trials each, 9~partial runs that we stopped early after 5 trials. We stopped a run early if the top 3 choices from CNN already suggested successful approach points, or if multiple successful grasps were executed in the first 5 trials. We obtained multiple successful task-oriented grasps for spatula, mug, hammers, pans, screwdrivers.

We could not complete BO on screwdriver handover task, knife and pen, because these required getting very close to the surface of the table (within 1-3mm). BO repeatedly proposed such approaches, but the planning library rejected these as near-collisions.

\begin{wrapfigure}{r}{0.35\textwidth}
\vspace{-15px}
\includegraphics[width=49mm]{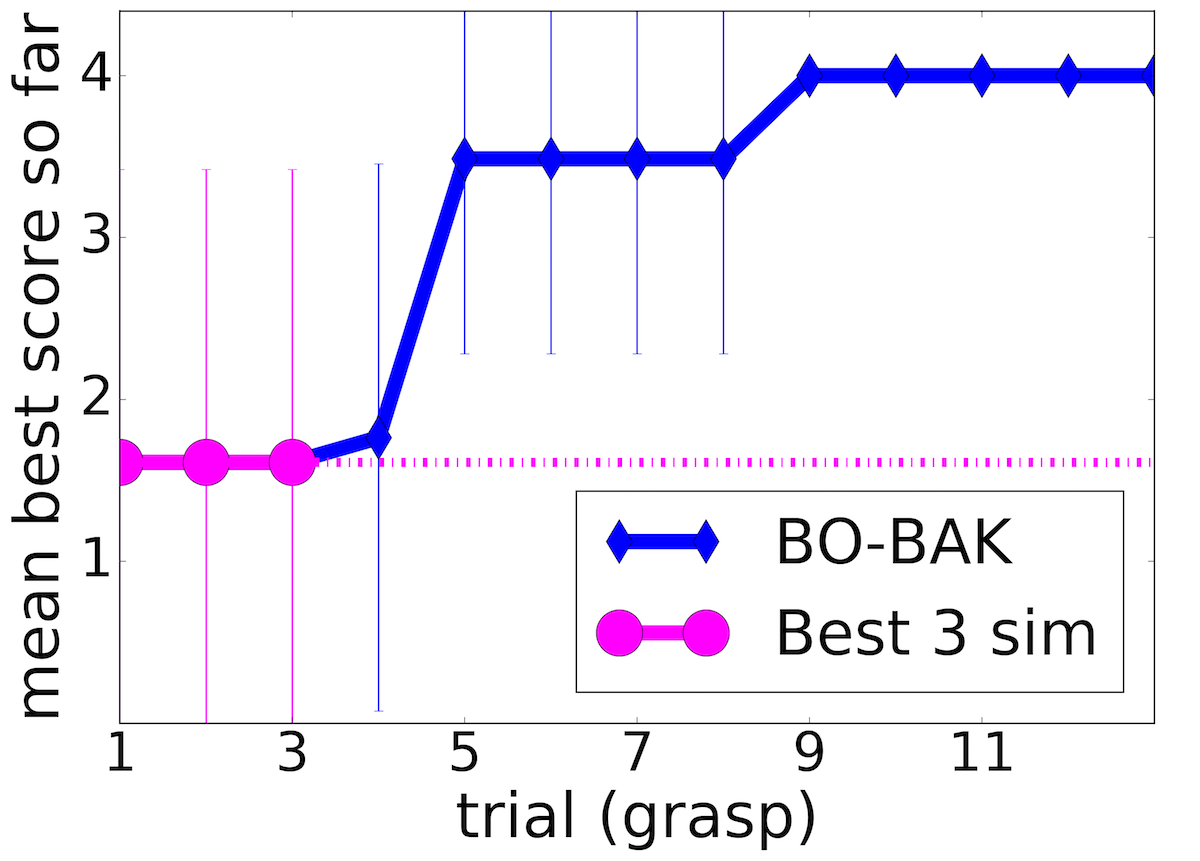}
\caption{\small{BO for challenging objects.}}
\label{fig:bo_full_trials}
\vspace{-20px}
\end{wrapfigure}
For the full BO runs we focused on object-task pairs that were challenging for top-10 approach (e.g. ran spatula \emph{handover} twice to ensure repeated success, since top-10 approach failed for this task). Figure~\ref{fig:bo_full_trials} summarizes the results, showing quick significant improvement of BO over the initial top choices from the 3 stability metrics.
Qualitatively, the benefits we observed from using BO were: 1) exploring various parts of the object systematically and efficiently; 2) sampling a variety of successful controllers that further improve over a merely acceptable controller.

\vspace{-8px}
\section{Conclusion and Future Work}
\label{sec:conclusion}
\vspace{-6px}
We proposed a variant of online global search suitable for simulation-informed optimization. Our focus was on validating this approach on a challenging robotics task. We plan to extend evaluation to task-oriented grasping scenarios in clutter, external disturbances, or real-time requirements for task completion. Our data generation does not assume a tabletop scenario, since objects are simulated without a support surface. This is a strength, since there is no need to re-run offline training when the workspace properties change. However, we need to improve our planning pipeline to avoid needless planning failures when grasping very small objects from tabletop. On the theory side, it would be interesting to explore theoretical properties of the proposed informed BO, investigate which guarantees could be obtained. It would be useful to put an emphasis on retaining realistic assumptions: no bounds on simulation-reality mismatch a-priori, non-smooth objective/cost functions.



\clearpage
\acknowledgments{\thanks{This research was supported in part by the Knut and Alice Wallenberg Foundation.}}


\bibliography{references}  

\begin{thebibliography}{35}
\providecommand{\natexlab}[1]{#1}
\providecommand{\url}[1]{\texttt{#1}}
\expandafter\ifx\csname urlstyle\endcsname\relax
  \providecommand{\doi}[1]{doi: #1}\else
  \providecommand{\doi}{doi: \begingroup \urlstyle{rm}\Url}\fi

\bibitem[Shahriari et~al.(2016)Shahriari, Swersky, Wang, Adams, and
  de~Freitas]{BOtutorial2016}
B.~Shahriari, K.~Swersky, Z.~Wang, R.~P. Adams, and N.~de~Freitas.
\newblock {Taking the Human Out of the Loop: A Review of Bayesian
  Optimization}.
\newblock \emph{Proceedings of the IEEE}, 104\penalty0 (1):\penalty0 148--175,
  2016.

\bibitem[Cully et~al.(2015)Cully, Clune, Tarapore, and Mouret]{cully2015robots}
A.~Cully, J.~Clune, D.~Tarapore, and J.-B. Mouret.
\newblock Robots that can adapt like animals.
\newblock \emph{Nature}, 521\penalty0 (7553):\penalty0 503--507, 2015.

\bibitem[Rai et~al.(2018)Rai, Antonova, Song, Martin, Geyer, and
  Atkeson]{rai2018bayesian}
A.~Rai, R.~Antonova, S.~Song, W.~Martin, H.~Geyer, and C.~G. Atkeson.
\newblock {Bayesian Optimization Using Domain Knowledge on the ATRIAS Biped}.
\newblock In \emph{Robotics and Automation (ICRA), 2018 IEEE International
  Conference on}, 2018.

\bibitem[Marco et~al.(2017)Marco, Hennig, Schaal, and Trimpe]{marco2017lqr}
A.~Marco, P.~Hennig, S.~Schaal, and S.~Trimpe.
\newblock On the design of {LQR} kernels for efficient controller learning.
\newblock In \emph{56th {IEEE} Annual Conference on Decision and Control,
  {CDC}}, 2017.

\bibitem[Antonova et~al.(2017)Antonova, Rai, and Atkeson]{antonova2017deep}
R.~Antonova, A.~Rai, and C.~G. Atkeson.
\newblock Deep kernels for optimizing locomotion controllers.
\newblock In \emph{Conference on Robot Learning}, pages 47--56, 2017.

\bibitem[Rasmussen and Williams(2005)]{GPsMLBook}
C.~E. Rasmussen and C.~K.~I. Williams.
\newblock \emph{{Gaussian Processes for Machine Learning (Adaptive Computation
  and Machine Learning)}}.
\newblock The MIT Press, 2005.
\newblock ISBN 026218253X.

\bibitem[Rai et~al.(2018)Rai, Antonova, Meier, and Atkeson]{rai2018using}
A.~Rai, R.~Antonova, F.~Meier, and C.~G. Atkeson.
\newblock Using simulation to improve sample-efficiency of bayesian
  optimization for bipedal robots.
\newblock \emph{arXiv preprint arXiv:1805.02732}, 2018.

\bibitem[Diankov and Kuffner(2008)]{diankov2008openrave}
R.~Diankov and J.~Kuffner.
\newblock Openrave: A planning architecture for autonomous robotics.
\newblock \emph{Robotics Institute, Pittsburgh, PA, Tech. Rep.
  CMU-RI-TR-08-34}, 79, 2008.

\bibitem[Lenz et~al.(2015)Lenz, Lee, and Saxena]{lenz2015deep}
I.~Lenz, H.~Lee, and A.~Saxena.
\newblock Deep learning for detecting robotic grasps.
\newblock \emph{The International Journal of Robotics Research}, 34\penalty0
  (4-5):\penalty0 705--724, 2015.

\bibitem[Mahler et~al.(2017)Mahler, Liang, Niyaz, Laskey, Doan, Liu, Ojea, and
  Goldberg]{mahler2017dex}
J.~Mahler, J.~Liang, S.~Niyaz, M.~Laskey, R.~Doan, X.~Liu, J.~A. Ojea, and
  K.~Goldberg.
\newblock Dex-net 2.0: Deep learning to plan robust grasps with synthetic point
  clouds and analytic grasp metrics.
\newblock \emph{arXiv preprint arXiv:1703.09312}, 2017.

\bibitem[Schmidt et~al.(2018)Schmidt, Vahrenkamp, W{\"a}chter, and
  Asfour]{schmidt2018grasping}
P.~Schmidt, N.~Vahrenkamp, M.~W{\"a}chter, and T.~Asfour.
\newblock Grasping of unknown objects using deep convolutional neural networks
  based on depth images.
\newblock In \emph{2018 IEEE International Conference on Robotics and
  Automation (ICRA)}, 2018.

\bibitem[Kroemer et~al.(2010)Kroemer, Detry, Piater, and
  Peters]{kroemer2010combining}
O.~Kroemer, R.~Detry, J.~Piater, and J.~Peters.
\newblock Combining active learning and reactive control for robot grasping.
\newblock \emph{Robotics and Autonomous systems}, 58\penalty0 (9):\penalty0
  1105--1116, 2010.

\bibitem[Montesano and Lopes(2012)]{montesano2012active}
L.~Montesano and M.~Lopes.
\newblock Active learning of visual descriptors for grasping using
  non-parametric smoothed beta distributions.
\newblock \emph{Robotics and Autonomous Systems}, 60\penalty0 (3):\penalty0
  452--462, 2012.

\bibitem[Oberlin and Tellex(2018)]{oberlin2018autonomously}
J.~Oberlin and S.~Tellex.
\newblock Autonomously acquiring instance-based object models from experience.
\newblock In \emph{Robotics Research}, pages 73--90. Springer, 2018.

\bibitem[Mahler et~al.(2016)Mahler, Pokorny, Hou, Roderick, Laskey, Aubry,
  Kohlhoff, Kr{\"o}ger, Kuffner, and Goldberg]{mahler2016dex}
J.~Mahler, F.~T. Pokorny, B.~Hou, M.~Roderick, M.~Laskey, M.~Aubry,
  K.~Kohlhoff, T.~Kr{\"o}ger, J.~Kuffner, and K.~Goldberg.
\newblock Dex-net 1.0: A cloud-based network of 3d objects for robust grasp
  planning using a multi-armed bandit model with correlated rewards.
\newblock In \emph{Robotics and Automation (ICRA), 2016 IEEE International
  Conference on}, pages 1957--1964. IEEE, 2016.

\bibitem[Song et~al.(2015)Song, Ek, Huebner, and Kragic]{song2015task}
D.~Song, C.~H. Ek, K.~Huebner, and D.~Kragic.
\newblock Task-based robot grasp planning using probabilistic inference.
\newblock \emph{IEEE transactions on robotics}, 31\penalty0 (3):\penalty0
  546--561, 2015.

\bibitem[Antanas et~al.(2014)Antanas, Moreno, Neumann, de~Figueiredo, Kersting,
  Santos-Victor, and De~Raedt]{antanas2014high}
L.~Antanas, P.~Moreno, M.~Neumann, R.~P. de~Figueiredo, K.~Kersting,
  J.~Santos-Victor, and L.~De~Raedt.
\newblock High-level reasoning and low-level learning for grasping: A
  probabilistic logic pipeline.
\newblock \emph{arXiv preprint arXiv:1411.1108}, 2014.

\bibitem[Kokic et~al.(2017)Kokic, Stork, Haustein, and
  Kragic]{kokic2017affordance}
M.~Kokic, J.~A. Stork, J.~A. Haustein, and D.~Kragic.
\newblock Affordance detection for task-specific grasping using deep learning.
\newblock In \emph{Humanoid Robotics (Humanoids), 2017 IEEE-RAS 17th
  International Conference on}, pages 91--98. IEEE, 2017.

\bibitem[Detry et~al.(2017)Detry, Papon, and Matthies]{detry2017taskoriented}
R.~Detry, J.~Papon, and L.~Matthies.
\newblock Taskoriented grasping with semantic and geometric scene
  understanding.
\newblock In \emph{IEEE/RSJ International Conference on Intelligent Robots and
  Systems}, 2017.

\bibitem[Fang et~al.(2018)Fang, Zhu, Garg, Kurenkov, Mehta, Fei-Fei, and
  Savarese]{fang2018learning}
K.~Fang, Y.~Zhu, A.~Garg, A.~Kurenkov, V.~Mehta, L.~Fei-Fei, and S.~Savarese.
\newblock Learning task-oriented grasping for tool manipulation from simulated
  self-supervision.
\newblock \emph{arXiv preprint arXiv:1806.09266}, 2018.

\bibitem[Chang et~al.(2015)Chang, Funkhouser, Guibas, Hanrahan, Huang, Li,
  Savarese, Savva, Song, Su, et~al.]{chang2015shapenet}
A.~X. Chang, T.~Funkhouser, L.~Guibas, P.~Hanrahan, Q.~Huang, Z.~Li,
  S.~Savarese, M.~Savva, S.~Song, H.~Su, et~al.
\newblock Shapenet: An information-rich 3d model repository.
\newblock \emph{arXiv preprint arXiv:1512.03012}, 2015.

\bibitem[Wu et~al.(2015)Wu, Song, Khosla, Yu, Zhang, Tang, and Xiao]{wu20153d}
Z.~Wu, S.~Song, A.~Khosla, F.~Yu, L.~Zhang, X.~Tang, and J.~Xiao.
\newblock 3d shapenets: A deep representation for volumetric shapes.
\newblock In \emph{Proceedings of the IEEE conference on computer vision and
  pattern recognition}, pages 1912--1920, 2015.

\bibitem[Rubert et~al.(2017)Rubert, Kappler, Morales, Schaal, and
  Bohg]{rubert2017relevance}
C.~Rubert, D.~Kappler, A.~Morales, S.~Schaal, and J.~Bohg.
\newblock On the relevance of grasp metrics for predicting grasp success.
\newblock In \emph{Intelligent Robots and Systems (IROS), 2017 IEEE/RSJ
  International Conference on}, pages 265--272. IEEE, 2017.

\bibitem[Kim et~al.(2001)Kim, Oh, Yi, and Suh]{kim2001optimal}
B.-H. Kim, S.-R. Oh, B.-J. Yi, and I.~H. Suh.
\newblock Optimal grasping based on non-dimensionalized performance indices.
\newblock In \emph{Intelligent Robots and Systems, 2001. Proceedings. 2001
  IEEE/RSJ International Conference on}, volume~2, pages 949--956. IEEE, 2001.

\bibitem[Chinellato et~al.(2005)Chinellato, Morales, Fisher, and
  Del~Pobil]{chinellato2005visual}
E.~Chinellato, A.~Morales, R.~B. Fisher, and A.~P. Del~Pobil.
\newblock Visual quality measures for characterizing planar robot grasps.
\newblock \emph{IEEE Transactions on Systems, Man, and Cybernetics, Part C
  (Applications and Reviews)}, 35\penalty0 (1):\penalty0 30--41, 2005.

\bibitem[Ponce and Faverjon(1995)]{ponce1995computing}
J.~Ponce and B.~Faverjon.
\newblock On computing three-finger force-closure grasps of polygonal objects.
\newblock \emph{IEEE Transactions on robotics and automation}, 11\penalty0
  (6):\penalty0 868--881, 1995.

\bibitem[Ponce et~al.(1997)Ponce, Sullivan, Sudsang, Boissonnat, and
  Merlet]{ponce1997computing}
J.~Ponce, S.~Sullivan, A.~Sudsang, J.-D. Boissonnat, and J.-P. Merlet.
\newblock On computing four-finger equilibrium and force-closure grasps of
  polyhedral objects.
\newblock \emph{The International Journal of Robotics Research}, 16\penalty0
  (1):\penalty0 11--35, 1997.

\bibitem[Ding et~al.(2001)Ding, Lee, and Wang]{ding2001computation}
D.~Ding, Y.-H. Lee, and S.~Wang.
\newblock Computation of 3-d form-closure grasps.
\newblock \emph{IEEE Transactions on Robotics and Automation}, 17\penalty0
  (4):\penalty0 515--522, 2001.

\bibitem[Cornella and Su{\'a}rez(2005)]{cornella2005fast}
J.~Cornella and R.~Su{\'a}rez.
\newblock Fast and flexible determination of force-closure independent regions
  to grasp polygonal objects.
\newblock In \emph{Robotics and Automation, 2005. ICRA 2005. Proceedings of the
  2005 IEEE International Conference on}, pages 766--771. IEEE, 2005.

\bibitem[Liu(2000)]{liu2000computing}
Y.-H. Liu.
\newblock Computing n-finger form-closure grasps on polygonal objects.
\newblock \emph{The International journal of robotics research}, 19\penalty0
  (2):\penalty0 149--158, 2000.

\bibitem[Li et~al.(2002)Li, Yu, and Tsujio]{li2002analytical}
Y.~Li, Y.~Yu, and S.~Tsujio.
\newblock An analytical grasp planning on given object with multifingered hand.
\newblock In \emph{Robotics and Automation, 2002. Proceedings. ICRA'02. IEEE
  International Conference on}, volume~4, pages 3749--3754. IEEE, 2002.

\bibitem[Surjanovic and Bingham(2013)]{surjanovic2013virtual}
S.~Surjanovic and D.~Bingham.
\newblock Virtual library of simulation experiments: test functions and
  datasets.
\newblock \emph{Simon Fraser University, Burnaby, BC, Canada}, 13:\penalty0
  2015, 2013.

\bibitem[Varley et~al.(2017)Varley, DeChant, Richardson, Ruales, and
  Allen]{varley2017shape}
J.~Varley, C.~DeChant, A.~Richardson, J.~Ruales, and P.~Allen.
\newblock Shape completion enabled robotic grasping.
\newblock In \emph{Intelligent Robots and Systems (IROS), 2017 IEEE/RSJ
  International Conference on}, pages 2442--2447. IEEE, 2017.

\bibitem[Rasmussen and Nickisch(2010)]{GPMLCode}
C.~E. Rasmussen and H.~Nickisch.
\newblock Gaussian processes for machine learning (gpml) toolbox.
\newblock \emph{J. Mach. Learn. Res.}, 11:\penalty0 3011--3015, Dec. 2010.

\bibitem[Gardner et~al.(2014)Gardner, Kusner, Xu, Weinberger, and
  Cunningham]{gardner2014bayesian}
J.~R. Gardner, M.~J. Kusner, Z.~E. Xu, K.~Q. Weinberger, and J.~Cunningham.
\newblock {Bayesian Optimization with Inequality Constraints.}
\newblock In \emph{ICML}, pages 937--945, 2014.

\end{thebibliography}

\end{document}